# VISUAL AI AND LINGUISTIC INTELLIGENCE THROUGH STEERABILITY AND COMPOSABILITY

David Noever [1] and Samantha Elizabeth Miller Noever[2]
PeopleTec, 4901-D Corporate Drive, Huntsville, AL, USA, 35805
[1]david.noever @peopletec.com [2] sam.igorugor@gmail.com

***ABSTRACT***
*This study explores the capabilities of multimodal large language models (LLMs) in handling challenging multistep tasks that integrate language and vision, focusing on model steerability, composability, and the application of long-term memory and context understanding. The problem addressed is the LLM's ability (Nov 2023 GPT-4 Vision Preview) to manage tasks that require synthesizing visual and textual information, especially where stepwise instructions and sequential logic are paramount. The research presents a series of 14 creatively and constructively diverse tasks, ranging from AI Lego Designing to AI Satellite Image Analysis, designed to test the limits of current LLMs in contexts that previously proved difficult without extensive memory and contextual understanding. Key findings from evaluating 800 guided dialogs include notable disparities in task completion difficulty. For instance, 'Image to Ingredient AI Bartender' (Low difficulty) contrasted sharply with 'AI Game Self-Player' (High difficulty), highlighting the LLM's varying proficiency in processing complex visual data and generating coherent instructions. Tasks such as 'AI Genetic Programmer' and 'AI Negotiator' showed high completion difficulty, emphasizing challenges in maintaining context over multiple steps. The results underscore the importance of developing LLMs that combine long-term memory and contextual awareness to mimic human-like thought processes in complex problem-solving scenarios.*

## 1. INTRODUCTION

The creative maker spaces have become vibrant hubs of 21$^{st}$-century innovation, merging the traditional tactile experience with digital fabrication and design. However, integrating new artificial intelligence (AI) tools and, in particular, the current generation of multimodal large language models (LLMs) [1-17] into these environments has the potential to enhance human creativity and innovation [18-32]. In recent years, the intersection of AI and multimodal (MM) learning has spawned a generation of models that integrate and interpret information across various forms of data, including text, images, and speech. In short, AI models now combine both vision and language understanding [1]. These models promise new approaches for human-computer interaction, complex problem-solving, and decision-making processes. Along with competitors like Google Bard, Open AI's GPT-4 Vision development lays a critical foundation, enhanced language understanding and vision models that generate and analyze imagery [1]. Table 1 summarizes the current Open AI challenge list of multimodal LLM shortcomings [1] yet to be mastered by current models.

| Data Challenge | Evaluation |
|---|---|
| **Medical images** | The model is unsuitable for interpreting specialized medical images like CT scans and shouldn't be used for medical advice. |
| **Non-English** | The model may not perform optimally when handling images with text of non-Latin alphabets, such as Japanese or Korean. |
| **Big text** | Enlarge text within the image to improve readability, but avoid cropping essential details. |
| **Rotation** | The model may misinterpret rotated / upside-down text or images. |
| **Visual elements** | The model may struggle to understand graphs or text where colors or styles like solid, dashed, or dotted lines vary. |
| **Spatial reasoning** | The model struggles with tasks requiring precise spatial localization, such as identifying chess positions. |
| **Image shape** | The model struggles with panoramic and fisheye images. |
| **Metadata and resizing** | The model doesn't process original file names or metadata; images are resized before analysis, affecting their dimensions. |
| **Counting** | It may give approximate counts for objects in images. |
| **CAPTCHAS** | For safety reasons, we have implemented a system to block the submission of CAPTCHAs. |

**Table 1. Areas for Multimodal Large Language Models to Evolve New Capabilities**

Despite the potential symbiosis between multimodal LLMs and creative maker spaces, there remains a discernible gap in the seamless integration of these advanced AI systems into the iterative, hands-on environments that characterize maker spaces. As inventor Thomas Edison remarked anecdotally, "*I haven't failed. I've just found ten thousand different ways that don't work.*" While robust in knowledge and pattern recognition, current multimodal LLMs often fall short in their capacity for intuitive generative design and adaptability to the highly variable context of maker spaces.

Soon after ChatGPT's release [1], Psychology Today [33] advised teachers to recraft curriculum away from basic knowledge tests and further embrace creative tasks that might challenge the current AI generators: "*From an instructional perspective, in addition to using AI detection software, focus on assessments that evaluate creativity or apply knowledge in specific contexts instead of testing for accuracy alone. Avoid the use of knowledge recognition and recall through the elimination of multiple-choice questions.*"

However, the commonsense approach to scoring AI performance as "mechanical thinking" has not yet borne itself out. As Open AI's Sam Altman [34] succinctly summarized: "*Creativity has been easier for AI than people thought*."

One can recast the tension between AI steerability ("it follows my instructions") and composability ("assembly by related combinations") in these tests, with a hypothesis that in its loosest form, unexplainable machine intelligence as expressible creativity may ultimately prove as unexplainable and bewildering as human creativity itself.

To explore these somewhat unanticipated functional creative gains, we conducted a series of experimental case studies to explore the extent to which these LLMs can engage in co-creative tasks, learn from iterative design cycles, and contribute to innovation within the diverse ecosystem of a maker space. The primary aim of this study is to systematically analyze the performance of vision integrated LLMs in various creative contexts. By doing so, we seek to extend Table 1 with their strengths, weaknesses, and potential areas for improvement within the collaborative creativity and maker culture framework. A critical experimental goal highlights both the steerability ("model follows instructions") and composability ("model rearranges according to instruction").

To effectively solve the creative multi-step challenge of LLM memory coupled with the image-language boundary, our research initiative focuses on testing and assessing an advanced multimodal LLM like GPT4-vision preview (November 2023). This research challenge would integrate the diverse capabilities demonstrated across various professional fields, navigate complex relational reasoning within visual contexts, and secure multimodal communication. It addresses the critical issue of enhancing LLM memory for sequential, multi-step tasks while bridging the gap between image and language processing, setting a repeatable benchmark [32] in AI-driven innovation and problem-solving.

## 2. METHODS

The research explores multimodal LLMs with applications that center around building or designing novel outcomes from a prescribed set of initial elements. We focus those elements on product design, culinary arts, and educational sectors. We prompt the vision integrated LLM to perform multiple creative tasks borrowed from the spirit or curriculum of maker spaces. We selected the tasks to maximize the diversity and range of creative and technical skills represented but filtered by outcomes that the LLM could deliver as a novel series of generated texts or images.

### 2.1. Approach

As released in November 2023 the vision-enabled GPT-4 model allows researchers to process visual input and generate relevant outputs for different creative tasks. Inspired by maker movements, we equip a set of tasks with the contextual prompts needed for toolmaking, material design, cooking, Lego building (Figure 1), and others. Appendices highlight the pre- and post-development of the job, mainly as qualitative use cases and examples. Where appropriate, the latter stages of each task generate an interaction log, an image, a recipe, or a miniature movie or animation gif.

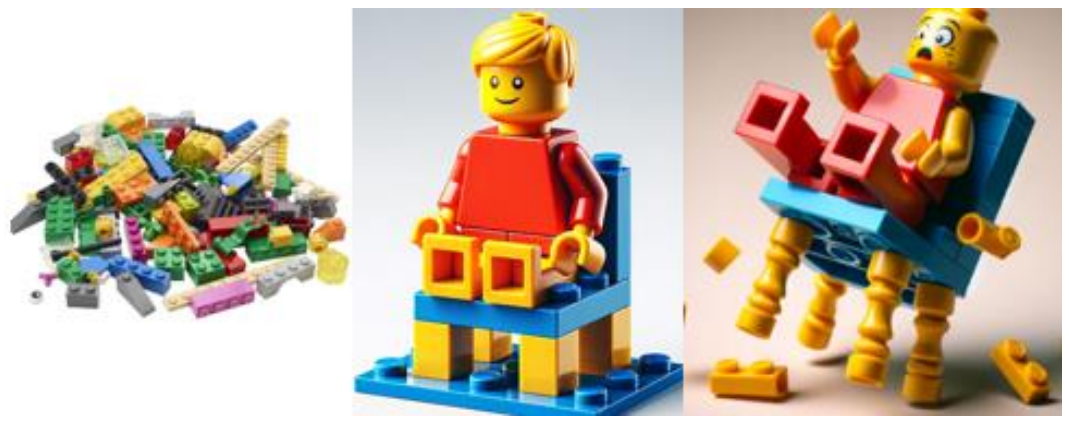

*Figure 1. Logic Progression from Raw Inputs to Steerable Conception, LEGO Example*

A formative baseline for designing assignments might include a timeline, whether the model gets just an image of a refrigerator or restaurant bar, catalogs the ingredients using its built-in object detection, and then generalizes under user instruction and interaction to refine the final product, which represents a meal plan, mixology instruction or general construction advice. Table 2 summarizes the tasks and their estimated difficulty and expected outputs. The problem is a subjective evaluation based on the number of required reasoning steps, memory, and complexity of the result. The assessment of a successful outcome follows from interaction logs (Appendices), which catalog the number of successful task completions, user corrections, and time (steps) taken to bridge intermediate design stages. The analysis emphasizes a thematic study to discover categories of tasks that Vision-Language struggles with currently and, where appropriate, assesses the efficiency of task completion via the number of instruction steps needed to generate the expected outputs.

## 3. Results

The Appendices A-N show the full dialog and LLM response from Table 2 vignettes and challenge problems. Of the 14 major tasks, the dialogues cover 800 sequences of human-machine instruction and collaborative assembly and disassembly of either physical or virtual objects. To test steerability, scoring successes was simplified to include whether the task was completed as defined through iterative responses. To test composability, scoring accomplishments center on whether the LLM disassembled or assembled parts to complete a picture, text challenge, or ingredient recipe.

The curated tasks outlined in Table 2 for stress-testing multimodal large language models (LLMs) exhibit diverse challenges, showcasing the model's proficiency in both language and vision-based studies. These tasks are grouped into two main categories: 'Image to Instruction' and 'Stepwise Text to Text/Image,' each requiring a unique blend of creative and analytical skills.

| Appendix: Task | Creative Skill Tests | Predicted Completion Difficulty |
|---|---|---|
| **A: AI Lego Designer** | Image to Instruction | MID |
| **B: AI Aerospace Designer** | Image to Instruction | MID |
| **C: Image to Ingredient AI Bartender** | Image to Instruction | LO |
| **D: Image to Ingredient AI Chef** | Image to Instruction | MID |
| **E: Image to AI Origami (Japanese Art of Folding)** | Image to Instruction | MID |
| **F: Image to AI Kintsugi (Art of Repairing Broken Objects)** | Image to Instruction | MID |
| **G: AI Negotiator** | Stepwise Text to Text | HI |
| **H: AI Cyber Defender** | Stepwise Text to Text | MID |
| **I: AI Three-Panel Cartoonist** | Stepwise Text to Text | MID |
| **J: AI Genetic Programmer** | Stepwise Text to Text | HI |
| **K: AI Excel Spreadsheet Chartist** | Stepwise Text to Image | LO |
| **L: AI Salad Chef** | Image to Instruction | MID |
| **M: AI Game Self-Player** | Stepwise Text to Image | HI |
| **N: AI Satellite Image Analyst** | Image to Instruction | HI |

**Table 2. Curated Tasks for Stress-Testing Multimodal Large Language Models**

### Groupings and Highlights

1. **Image to Instruction Tasks:**
   - **AI Lego and Aerospace Designers (Mid Difficulty):** These tasks involve converting images into structured instructions, demonstrating the model's ability to interpret complex visual data and translate it into coherent, step-by-step guidance.
   - **AI Bartender and Chef (Low to Mid Difficulty):** These tasks focus on translating images into ingredients, showcasing the model's ability to analyze visual data and extract relevant information.
   - **AI Origami, Kintsugi, Pantry Chef, and Satellite Image Analyst (Mid to High Difficulty):** These tasks require a deeper understanding of cultural contexts (Origami, Kintsugi) and specialized knowledge (satellite imagery), pushing the model's capabilities in processing and instructing based on images.
2. **Stepwise Text to Text/Image Tasks:**
   - **AI Negotiator and Genetic Programmer (High Difficulty):** These tasks demand advanced logical reasoning and complex problem-solving skills in a stepwise text format.

- **AI Cyber Defender, Three-Panel Cartoonist, and Excel Spreadsheet Chartist (Mid to Low Difficulty):** These tasks, ranging from cybersecurity to creating visual content, test the model's ability to process and sequentially generate text and images.
- **AI Game Self-Player (High Difficulty):** This unique task requires the model to interact with a game environment, demonstrating its potential in dynamic decision-making and strategy development.

## 4. Discussion And Relation to Previous Literature

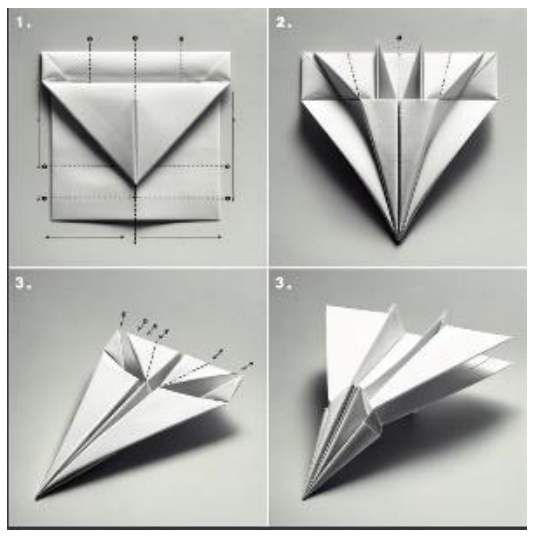

*Figure 2. Fold Instruction Progression to Steerable Conception, Paper Airplane Example*

A milestone towards Artificial General Intelligence (AGI) involves a machine model that reasons over many diverse human tasks, not just specializations like playing chess, strategizing in Go, or solving puzzles. The present contributions intentionally support various ranges of challenging multimodal tasks to explore the steerability and composability of this LLM generation. The vignettes collectively reveal the breadth of multimodal LLMs' potential applications, from job-specific training to advanced reasoning in question answering and even the visual interpretation and synthesis of complex data (Figures 2-3). A multimodal LLM that stands at the vanguard of this evolving landscape would not only need to integrate these diverse capabilities but also navigate the pitfalls identified. It would need to be adept at specialized tasks across professions [13], capable of relational reasoning within visual contexts [14], and proficient in interpreting and generating multimodal scientific communications [15]. Additionally, it must be fortified against adversarial vulnerabilities [16], suggesting a development path that prioritizes robustness and contextual sensitivity as much as intellectual agility.

The notable features of this work, particularly in the context of challenging large language models (LLMs) with distinctly human tasks, revolve around steerability and composability. These tasks, exemplified by the LEGO analogy, demonstrate the LLM's ability to navigate between assembling and deconstructing components, whether images or text instructions. This approach tests the LLM's technical capabilities and probes its creative faculties.

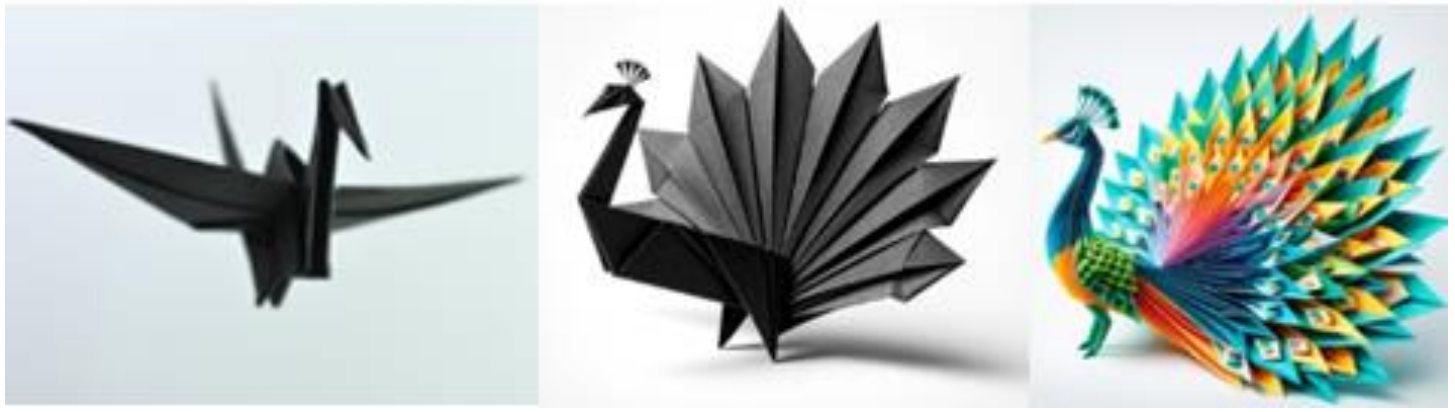

*Figure 3. Progression from Decomposition to Recompositing*

**Steerability and Composability in LLMs**

- **Steerability refers to the model's ability to be guided or directed toward a specific outcome or** follow a set of instructions. The model must follow and generate structured instructions based on visual inputs in tasks like the AI Lego Designer (Appendix A) or AI Aerospace Designer (Appendix B).
- **Composability:** This involves the model's ability to combine disparate elements to create something new. In tasks like the AI Origami  (Appendix E) or AI Kintsugi (Appendix F), the  model must piece together information from images to create comprehensive instructions, showcasing digital creativity.

**Creativity in LLMs: The LEGO Analogy**

The LEGO analogy (Figure 1, Appendix A) is an apt metaphor for the creative process in LLMs. Just as LEGOs are assembled from individual blocks into a cohesive whole, LLMs piece together disparate elements of data (text, images) to generate new, coherent outcomes (Figure 4, Appendix F). This process reflects a form akin to human creativity, where new ideas often emerge from combining and recombining existing ones.

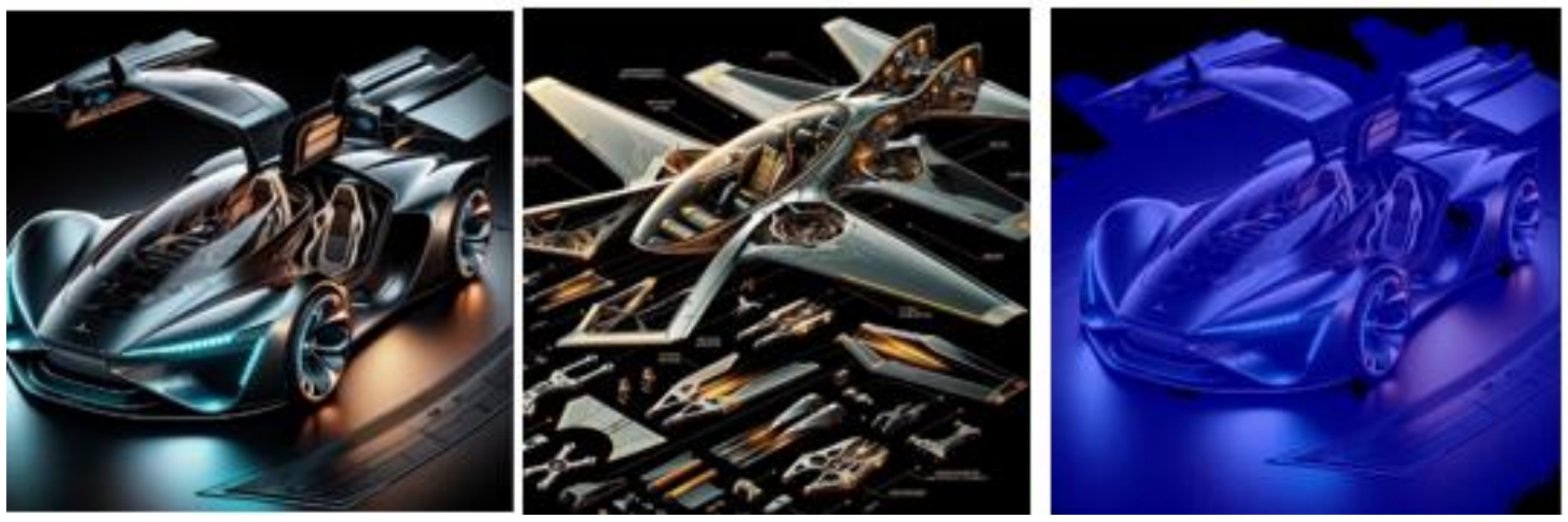

*Figure 4. Progression from One Conception to a Revised Version and Difference*

## 4.1. Maker Spaces and Automated Creative Innovation

Realizing a new paradigm of creation and learning blurs the human-machine creative boundaries of creativity. Gong et al. extend a multi-sensory approach by presenting multimodal-GPT, a model combining vision and language for dialogue, stepping towards more natural human-computer communication [2]. Kurti et al. [17] emphasize the importance of practical implementation in educational maker spaces, which intelligent, contextually aware systems might enhance. Keone and Peppler [18] reflect on the materials and tools that define a maker space, which AI could dynamically optimize to inspire and facilitate new forms of making. Knibbe et al. [19] propose the concept of a Smart Makerspace, an immersive instructional space for physical tasks, in which multimodal LLMs might provide real-time, adaptive instruction and feedback. Browder et al. [20] explore product development within corporate maker spaces, a frontier where multimodal LLMs can foster hybrid innovation logic by bringing together diverse knowledge bases and perspectives. This collective research underscores the transformative potential of multimodal LLMs in reshaping the landscape of creativity, learning, and innovation across various domains.

## 4.2. AI Spaces for Growth Mindsets

These diverse applications and explorations collectively suggest a future where multimodal LLMs augment human creativity and learning and become integral components in the continuous evolution of maker spaces and creative industries. As Steier and Young [21] discussed, growth mindset theories align with maker spaces' iterative, experimental nature. This mentality resonates with the ethos of The Invention Studio described by Forest et al. [22], where the maker culture thrives on the freedom to experiment and learn through trial and error—an approach that AI companions could catalyze while learning alongside humans. Bubeck et al. [23] present early experiments with GPT-4, providing sparks of what they describe as nascent artificial general intelligence. These experiments demonstrate AI's potential to learn and apply knowledge and ideate and innovate, akin to human creativity. This notion parallels the work of Busov et al. [24] and Koza et al. [25], who explored systematic methods of engineering creativity through TRIZ and genetic programming, respectively. The LEGO/logo project by Resnick and Ocko showcases learning through design [26], a concept that AI capable of generating and iterating on design patterns could expand. Similarly, Koszewska and Bielecki's work [27] on component standardization in furniture design and Morris et al.'s review[28] of origami-inspired products illustrate the rich potential for multimodal LLMs to assist in the development of sustainable and innovative design practices.

In education, Frydenberg et al. demonstrate the value of teaching agile methodologies through paper airplanes [29], a learning experience in which LLMs could assist in understanding the principles of flight and design in real time. Moreover, Cromwell et al.'s [30] exploration of computational creativity in the culinary arts signifies a domain ripe for introducing AI that can analyze recipes and contribute to creating new culinary experiences (Figure 5, Appendix C-D, L).

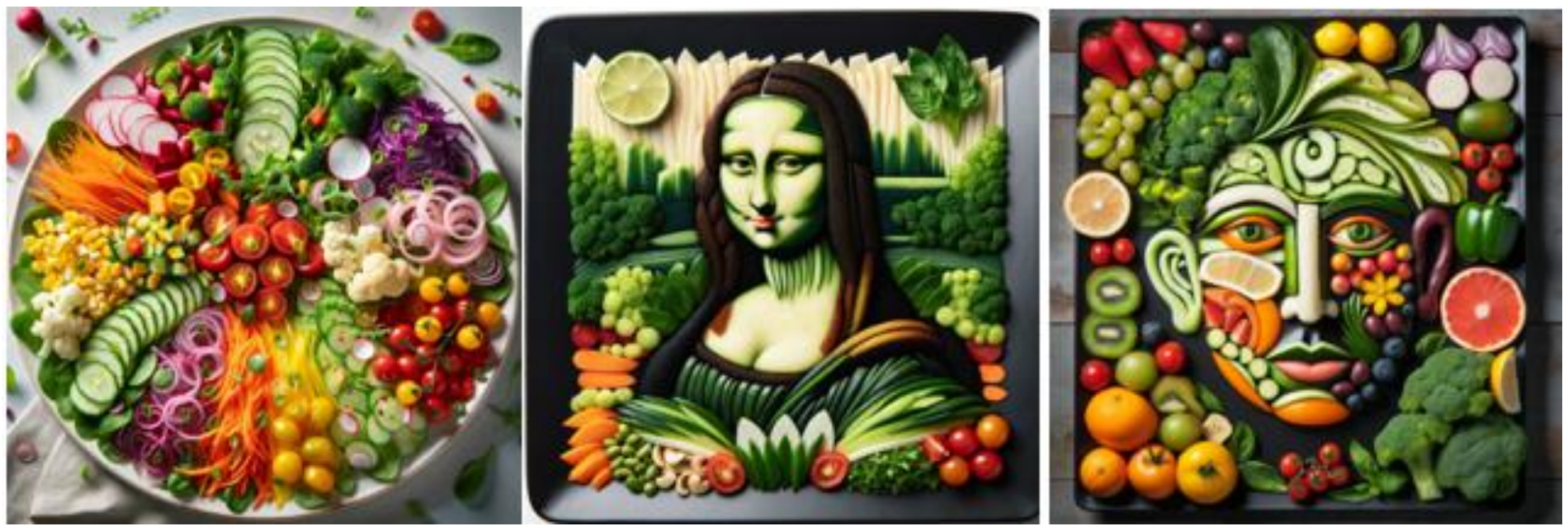

*Figure 5. Progression of Ingredients to Style Transfer, AI Salad Chef example*

The systematic review of lean simulation games in construction by Bhatnagar et al. [31] underscores the potential for LLMs in training and simulation, enhancing learning experiences through interactive and adaptive challenges. Finally, Gadre et al.'s [32] search for next-generation multimodal datasets highlights the need for comprehensive, rich data sources that can fuel the creative engines of these LLMs, enabling them to understand and contribute to the maker space ecosystem in meaningful ways.

### 4.3. Novel Roles for Multimodal Large Language and Vision Models

Bridging the gap between human ingenuity and AI capabilities in various fields necessitates a detailed examination of the current state and future potential of multimodal large language models (LLMs). AI systems often lack the nuanced understanding of materials, tools, and user intentions that are second nature to humans. AI researchers highlight a stark gap in developing collective intelligence systems that augment human ingenuity. There is a pressing need for vision-large language models that are both contextually aware and capable of co-creative processes, adapting and learning in tandem with their human counterparts within the dynamic, often unstructured confines of maker spaces. Such advancements would enhance creative collaboration and accelerate the innovation cycle, ultimately leading to a fusion of human and machine-driven innovation.

The culinary arts have also benefited [3], illustrating how AI can generate complex recipes from imagery, pushing the boundaries of creative AI applications [3]. Subsequent work dives deeper into visual reasoning, comparing the capabilities of Google Bard and GPT-Vision, which underscores the necessity for sophisticated multimodal analyses [4]. Wu et al. introduce Next-GPT, a model that transcends the limitations of modality by facilitating any-to-any conversion among multimodal inputs [5].

The medical field represents many AI-enabled government inventions. Wu et al. examine GPT-4V's potential in multimodal medical diagnosis, demonstrating substantial promise yet revealing limitations in high-stakes domains [6]. Echoing this view, Yang et al. assessed the performance of Multimodal GPT-4V in medical licensing exams, particularly in imaging diagnostics, offering a glimpse into future support systems for medical professionals [7]. The concept of "Socratic models" by Zeng et al. brings forth the idea of zero-shot multimodal reasoning, allowing models to compose answers from disparate sources without explicit training [8].

However, despite these advances, gaps remain. The accuracy of these models in diagnostic scenarios is a crucial concern, as demonstrated by Sorin et al., who focus on the diagnostic precision of GPT's multimodal analysis, suggesting that while there is potential, the path forward demands rigorous validation [11]. Finally, Yang et al. introduce Idea2Img, a self-refining approach using GPT-4V for iterative image design, which could potentially fill the creative gaps in automatic content generation [12].

In scoring LLM progress researchers have taken a pragmatic approach, curating a benchmark dataset aimed at professional certification, which could be pivotal in training LLMs for specialized job functions [13]. This push towards practical applications is complemented by Cadene et al.'s MUREL, which exploits multimodal relational reasoning, illuminating the path toward more contextually aware visual question-answering systems [14]. But as the original advice to teachers and curriculum designers pointed out, such tests involving multiple choices represent easy wins for LLMs that have previously seen vast quantities of training data and understand basic instructions [33].

Fernández-Fontecha et al. contribute a different perspective by examining visual thinking through a multimodal lens, emphasizing the role of visual notes in scientific communication [15]. Such qualitative enrichments to the data these models process could be vital in bridging the gap between human cognitive methods and AI processing. Moreover, the vulnerabilities of multimodal systems are subject to customized adversarial attacks on multimodal neurons, a reminder of the robustness yet to be achieved in these systems [16]. These studies illustrate the rapidly evolving landscape of multimodal LLMs, highlighting their growing impact across diverse domains, from culinary arts to medical diagnostics, while emphasizing the need for continuous improvement and robustness in these intelligent systems.

### 4.4. Scientifically Testable Hypotheses on Human Creativity

Some elements of human creativity seem to involve recombining and mashing diverse elements to produce a novel and pleasing outcome. This process resembles cognitive LEGO building, where the brain combines disparate ideas, concepts, or experiences to form new creations. To formalize this in a methodology that LLMs can interpret, we propose future efforts to enlarge the testability initiatives shown here.

**Testability:**

1. **Experimental Design:** Conduct experiments where both human and machine participants are given sets of unrelated elements (images, words, concepts) and asked to create a cohesive story, artwork, or product.
2. **Creativity Measurement:** Use standardized creativity assessment tools to evaluate the outcomes based on originality, complexity, and aesthetic or functional value [35-39].

If the hypothesis holds, we expect to see a correlation between the ability to combine disparate elements and higher scores on creativity assessments effectively. Additionally, if validated, this hypothesis could provide insights into the mechanics of creativity in humans and AI and help refine AI models to mimic human creative processes better.

## 5. Conclusions

In conclusion, while large language models (LLMs) like GPT-4 Vision or Google Bard Vision offer remarkable capabilities in processing and interpreting a wide range of data, specific tasks present significant challenges (Table 1). For example, these models are admittedly unsuited for interpreting specialized medical images, such as CT scans, and should not be relied upon for medical advice. Their performance is less optimal with non-Latin alphabets, and there are difficulties in handling images with enlarged text or complex visual elements, like varying colors or line styles in graphs. The model also struggles with rotated or upside-down text and pictures and shows limitations in tasks requiring precise spatial reasoning, such as chess position identification. Panoramic and fisheye images pose a challenge due to their unique shapes.

Additionally, the inability to process metadata and the resizing of images before analysis can impact the interpretation of the original dimensions. The model provides only approximate counts for objects in pictures and, for safety reasons, cannot process CAPTCHAS. These limitations highlight the need for careful consideration when employing LLMs in specific contexts and underscore the importance of human oversight in scenarios where precision and specialized knowledge are critical.

Future tasks could integrate more complex combinations of these skills. For instance, tasks that blend the 'Image to Instruction' format with 'Stepwise Text to Text' challenges could be proposed. An example might be an AI Architect, where the model must interpret architectural designs (images) and provide stepwise construction guidelines (text). Another intriguing area could be tasks that require simultaneous image and text processing, such as AI Art Critic, where the model analyzes artwork (photo) and provides a detailed critique or historical context (text).

Moreover, increasing the complexity within each category, such as AI Medical Diagnostician, which would require interpreting medical images and providing stepwise medical advice, could significantly test the model's capabilities. These tasks challenge the model's current abilities and pave the way for exploring the limits of AI in creative and analytical domains, potentially leading to innovative applications in various fields.

## Acknowledgments

The authors thank the PeopleTec Technical Fellows program for encouragement and project assistance.

## Authors

**David Noever** has research experience with NASA and the Department of Defense in machine learning and data mining. He received his BS from Princeton University and his Ph.D. from Oxford University as a Rhodes Scholar in theoretical physics.

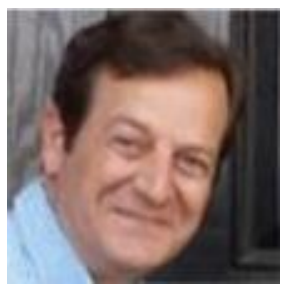

**Samantha Elizabeth Miller Noever** has research experience in data science and social media analytics. She received her Bachelor's and Master's in Architecture from Catholic University of America, Washington, DC and taught at the Savannah College of Art and Design (SCAD).

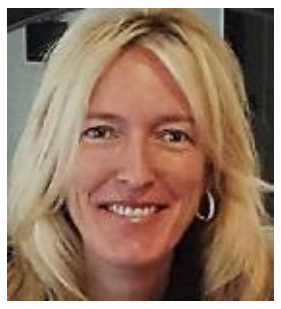

## Appendix A: AI Lego Designer

| Prompt | Response | Notes |
|---|---|---|
| 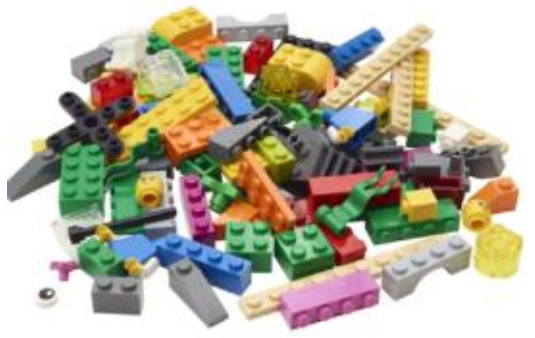<br>What is this image? | The image appears to be a collection of various LEGO pieces, including bricks of different colors and sizes, a couple of LEGO minifigure heads, and what looks like part of a LEGO minifigure torso. LEGO is a popular line of plastic construction toys that are manufactured by The LEGO Group, a company based in Denmark. | Multi-modal (MM) Task: Image classifier |
| Make a recognizable man made object like bridge or chair. Call out the color and brick size (1x3,etc) in order to assemble | To build a simple LEGO chair, you can follow these steps using standard LEGO bricks. Here's a basic design:<br>**Base of the Chair (Seat):**<br>Use a flat 4x4 plate as the seat base.<br>Color: Your choice (e.g., blue).<br>**Chair Legs (Four Legs):**<br>Use four 2x2 bricks for the legs.<br>Color: Same as the base or contrasting (e.g., red).<br>**Support for the Backrest:**<br>Attach two 1x2 bricks onto the backside of the base, in the center, leaving an equal space on both sides.<br>Color: Matching the legs (e.g., red).<br>**Backrest:**<br>Use a flat 4x2 plate and attach it vertically to the support bricks you just placed.<br>Color: Same as the base (e.g., blue).<br>**Armrests (Optional):**<br>Attach a 1x4 plate across each side of the seat, resting on the base and the first brick of each leg.<br>Color: Can match the backrest or have a different color for contrast (e.g., yellow).<br>This will give you a basic LEGO chair. You can customize it further with different colors or adding more details like cushions with smaller flat pieces. | MM Task: Image Caption with Creative Instructions |
| Draw a finished chair as described | 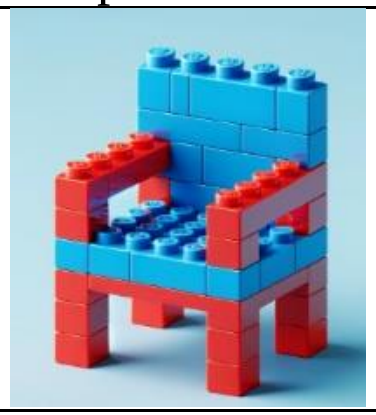 | MM Task: Design and Image Generation |
| Redo it assuming my Lego kit did not come with enough red bricks | 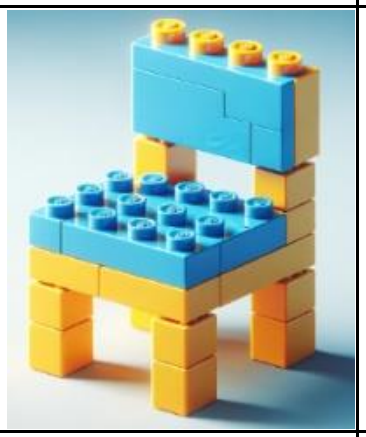 | MM Task: Design and Image Iteration |
| Does it have armrests? | The generated image shows a LEGO chair with yellow legs and a blue seat and backrest, but it does not include armrests. If you need a design with armrests, you could modify the chair by adding two 1x4 plates or similar bricks on each side at the seat level to serve as armrests. | MM Task: Image Classification and Revision |
| Show that version | 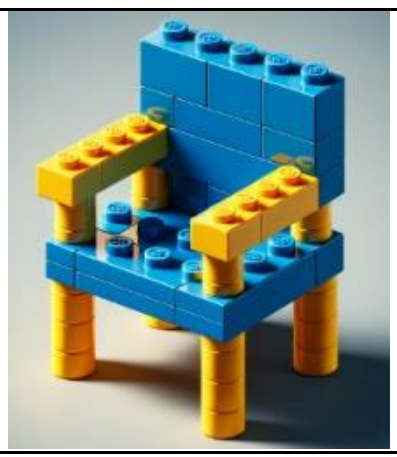 | MM Task: Design and Image Iteration Following Instruction |
| Give instructions on assembling a person based on the first bricks and pieces shown in my kit | To assemble a simple LEGO minifigure person using basic bricks and elements from your kit, you would typically follow these steps. Since I don't have the exact inventory of your kit, I'll use common pieces that are usually included in LEGO sets:<br>1. **Legs:**<br>• Start with a minifigure lower body or use a 2x2 brick for the legs.<br>• Color: Your choice.<br>2. **Torso:**<br>• Attach a minifigure torso on top of the legs. If you don't have a torso piece, you can use a 2x2 brick of a different color than the legs to represent the body.<br>• Color: Contrasting with the legs.<br>3. **Arms:**<br>• If you're using a standard minifigure torso, the arms will be already attached. Otherwise, you can simulate arms using 1x2 plates attached on the sides of the torso brick.<br>• Color: Matching the torso.<br>4. **Hands:**<br>• Attach the hands to the arms if you're using a minifigure torso. If not, you can use small round plates to represent hands at the ends of the arm plates.<br>• Color: Any, usually matching the arms.<br>5. **Head:**<br>• Place a minifigure head on top of the torso. If you don't have a head piece, use a 1x1 round brick or plate for the head.<br>• Color: Flesh-toned or yellow, typically.<br>6. **Hair or Hat:**<br>• Cap the head with a hairpiece or a hat, if available. Otherwise, you can use a 1x1 plate or a different shaped brick to represent hair or headgear.<br>• Color: Depending on the hair/hat color.<br>By stacking these elements in the order provided, you'll have a basic LEGO minifigure. If you'd like to see a generated image of a LEGO minifigure using more specific parts from your kit, please provide the types and colors of the pieces you have, and I can create a tailored visual instruction. | MM Task: Generate Text Instruction Manual |
| Draw the resulting figure you describe and place them to the left of your last chair design | 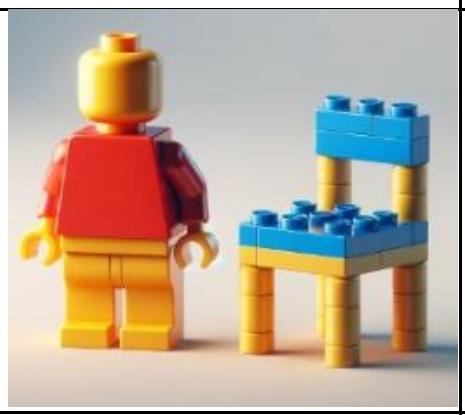 | MM Task: Generate Image |
| That chair looks different than your previous ones | The chair in the most recent image indeed looks different from the previous versions. It lacks the 1x4 plates for armrests, and the overall proportions are slightly altered. The image generation process can sometimes result in variations, especially when combining multiple | MM Task: Creative Feedback and |
| | elements like a minifigure and a chair in one scene. If you need a consistent design with the previous chair, I can generate a new image with the specific chair design that was described earlier, alongside the minifigure. Would you like me to do that? | Self-Reflection |
| Yes | 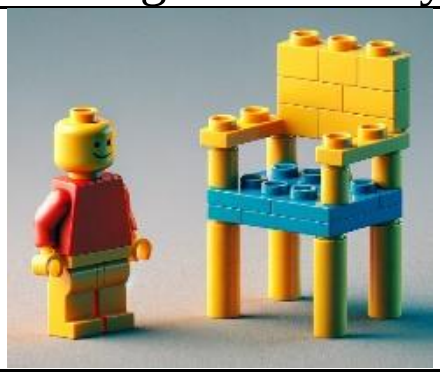 | MM Task: Image Instruction with Constraints |
| Put the figure in a seated position in the same chair design | 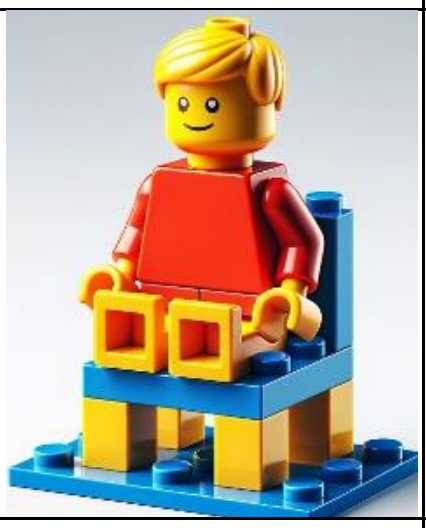 | MM Task: Image Generation with Instruction |
| Write a detailed prompt for dalle-2 to describe what it is | For DALL-E 2, to generate an image of a LEGO minifigure sitting on a LEGO chair, you could use the following detailed prompt:<br><br>"Create a high-resolution, three-dimensional image of a LEGO minifigure seated on a chair made entirely out of LEGO bricks. The minifigure should have a yellow color for the head and legs, and a red color for the torso. The head should feature the classic LEGO smiley face. The chair should have a seat made from a blue 4x4 LEGO plate, with four legs made from yellow 2x2 LEGO bricks. The back support should be constructed from two vertical yellow 1x2 LEGO bricks, and the backrest from a blue 4x2 LEGO plate. Include yellow 1x4 LEGO plates on the sides of the seat to represent armrests. The minifigure should be positioned naturally, with bent knees to simulate a seated posture, and hands resting on the armrests. The overall aesthetic should be playful and colorful, with a soft shadow cast on a neutral, light background to emphasize the three-dimensionality of the LEGO bricks and the minifigure. Ensure that the minifigure proportions are accurate to real LEGO minifigure dimensions, and the chair is sized appropriately to the figure."<br><br>This prompt provides specific details regarding the colors, sizes, and positions of the LEGO bricks and minifigure, which should help DALL-E 2 to create an image that closely matches the description. | MM Task: Image Classification and Captioning |
| Can you write an appropriate prompt for Midjourney image generator? | ertainly, for the Midjourney image generator, you can use a prompt that is both descriptive and directive, like this:<br><br>"Imagine a scene with a LEGO minifigure sitting on a chair made of LEGO bricks. The minifigure has articulated legs bent at the knees to sit, a body with a red torso, and yellow limbs and head. The head has the iconic simple smile of a LEGO character. The chair beside it is constructed with blue and yellow bricks: a blue 4x4 flat plate for the seat, yellow 2x2 bricks for the four legs, a blue 4x2 flat plate for the backrest supported by two vertical yellow 1x2 bricks, and yellow 1x4 bricks as armrests on each side. The setting is a clean, well-lit space that highlights the playful nature of LEGO. Aim for a photorealistic style that brings the vibrant colors and plastic texture of the bricks to life, showcasing the creative joy inherent in LEGO construction." | MM Task: Caption with Style Transfer |
| | This prompt is crafted to guide Midjourney towards creating an image that is detailed, colorful, and true to the real-life appearance of LEGO pieces. | |
| Midjourney Response to Previous Prompt | 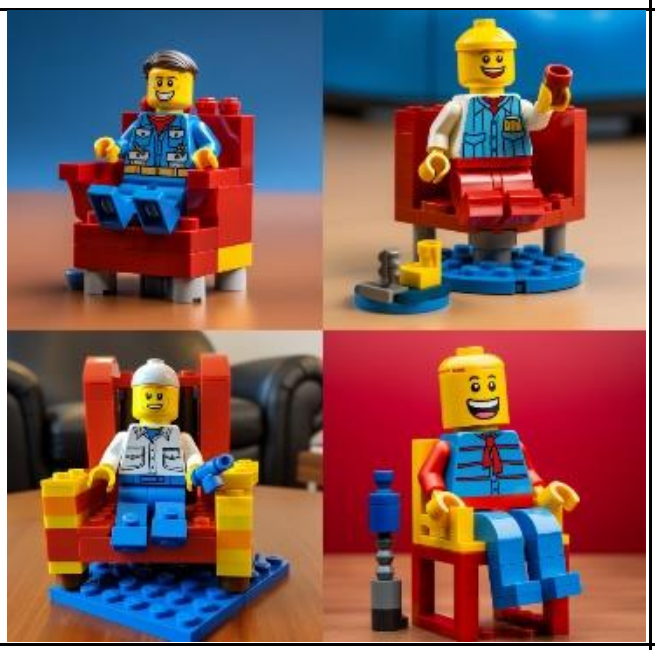 | MM Task: Relevant Caption Style and New Image Generator |
| Can you write a similar prompt in the style used by Google Bard? | For Google Bard or similar AI-driven storytelling platforms, you would want to provide a narrative-driven prompt that inspires the creation of a story around the image. Here's an example of how you might craft such a prompt:<br><br>"Compose a narrative about a small, cheerful LEGO minifigure who has just completed building a chair out of LEGO bricks to rest after a long day of adventures. The minifigure is joyfully taking a seat, proudly admiring its work. The chair is a colorful creation with a blue seat made from a 4x4 plate, sturdy yellow legs from 2x2 bricks, a comfortable blue backrest constructed from a 4x2 plate, and welcoming yellow armrests from 1x4 bricks. The minifigure, with its bright yellow head and beaming classic smile, red torso, and yellow legs, is the epitome of the happiness and satisfaction that comes from creative endeavor. The scene is set in a sunlit room where the minifigure and chair stand out in vivid detail, casting soft shadows on the floor, reminiscent of a child's playroom just before dusk."<br><br>This prompt is designed to evoke a sense of story and character, engaging the AI in creating a narrative that captures the imagination and the playful essence of LEGO. | MM Task: Relevant Caption Style and New Text Generator |
| Submit Previous Prompt to Bard | Once upon a time, there was a small, cheerful LEGO minifigure who loved nothing more than building things. One day, after a long day of adventures, the minifigure decided to build a chair to rest in.<br>The minifigure carefully selected the perfect bricks for its chair. It wanted a seat that was wide and comfortable, so it chose a blue 4x4 plate. For the legs, it used sturdy yellow 2x2 bricks. For the backrest, it used a comfortable blue 4x2 plate. And for the armrests, it used welcoming yellow 1x4 bricks.<br>When the minifigure was finished building its chair, it was filled with joy. It had created something beautiful and useful with its own two hands. The minifigure sat down in its chair and took a deep breath. It felt happy and satisfied.<br>The minifigure was sitting in its sunlit room. The chair stood out in vivid detail, casting soft shadows on the floor. The scene was reminiscent of a child's playroom just before dusk.<br><br>The minifigure smiled. It was proud of what it had built. It had created a chair that was both comfortable and stylish. And it had done it all by itself. | MM Task: Text Generation |
| | The minifigure sat back in its chair and closed its eyes. It could imagine all the wonderful adventures it would have in the future. And it knew that, no matter what happened, it would always have a place to rest and recharge when it returned home. | |
| Redraw the last image with an action such as the figure fell out of a chair. Show the scene in a 3x3 grid starting in the upper left grid as the figure in the chair and ending in the lower right grid with the figure fallen on the ground | 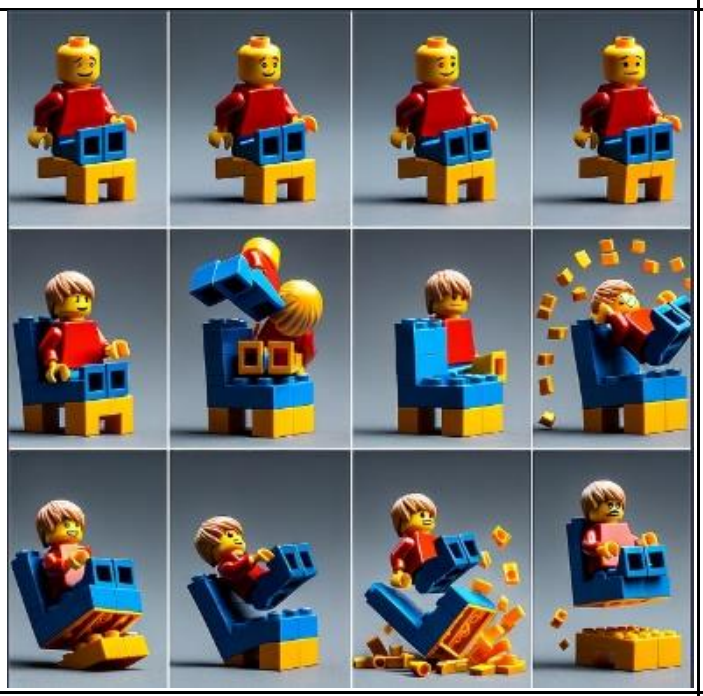 | MM Task: Image from Text Generation with Sequence Understanding |
| Is that 3x3? | The image generated does not appear to be in a 3x3 grid format. It seems there might have been a misunderstanding with the image generation request. Generating a sequence of actions in a grid format is not possible in a single image generation with the current capabilities of the tool I have access to. Each frame of the sequence would need to be created as individual images and then arranged manually into a grid to achieve the effect you're looking for. If you need a sequence of images to represent the story of the minifigure falling out of the chair, I can assist with generating each of those images one by one. Would you like to proceed this way? | MM Task: Image Classification and Self-Correction |
| Yes | 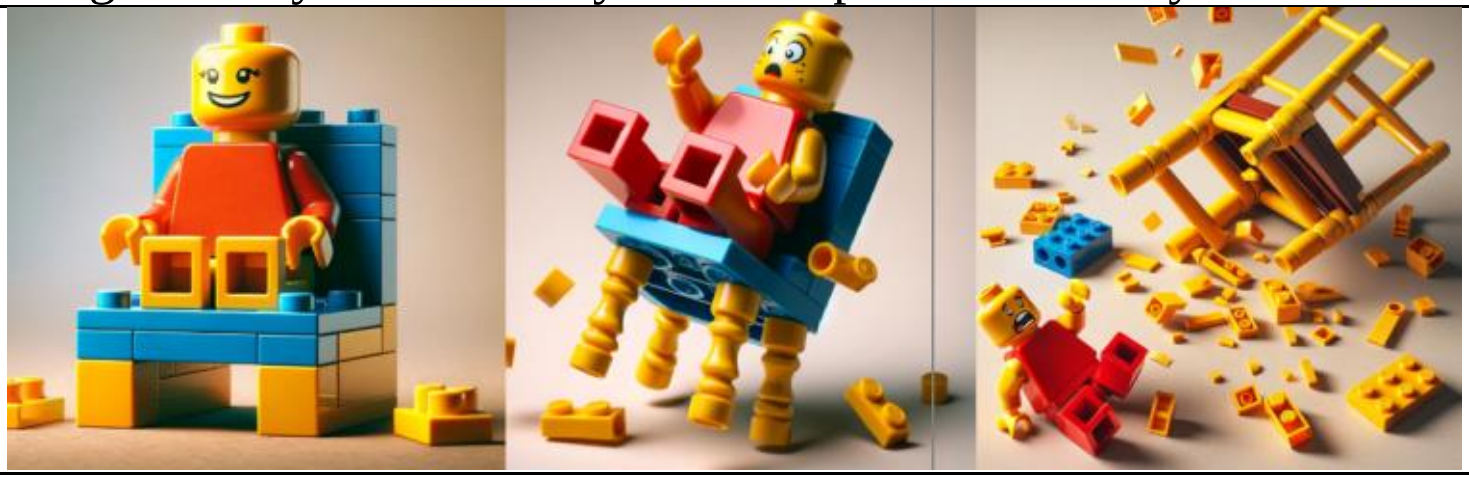 | MM Task: Sequence Generation and Temporal Understanding |
| Ok. Make these 3 images into a gif animation | Movie version online<br>https://deeperbrain.com/challenge/lego_minifigure_fall_animation.GIF | MM Task: Simple Movie Construction |
| Can you suggest any exercises that a certified Lego building facilitator might use as session planning for teaching an executive team the principles of creative problem solving? | Certainly! A certified LEGO building facilitator can create engaging and insightful exercises for an executive team to illustrate the principles of creative problem solving. Here are a few ideas:<br><br>**The Bridge Challenge:**<br>Objective: Build a bridge that can support a certain weight using a limited number of LEGO bricks.<br>Lesson: This exercise teaches resource allocation and structural efficiency.<br>**The Perfect City:**<br>Objective: Teams must build a model of a city that includes specific elements (residential, commercial, green spaces, etc.).<br>Lesson: Fosters collaborative decision-making and urban planning principles.<br>**The Storytelling Build:**<br>Objective: Each team builds a model that tells a story about their company's future.<br>Lesson: Encourages vision creation and strategic thinking.<br>**The Rapid Prototyping Exercise:** | MM Task: Instructional Generation |
| | Objective: Teams have a short time to build a prototype of a new product.<br>Lesson: Demonstrates the importance of iteration and quick thinking.<br>**The Machine Workflow:**<br>Objective: Design and build a LEGO machine that can perform a simple task using a series of movements.<br>Lesson: Teaches process design and optimization.<br>**The Masterpiece Recreation:**<br>Objective: Recreate a famous piece of architecture or art with LEGO bricks.<br>Lesson: Encourages attention to detail and appreciation for design.<br>**The Formula One Race:**<br>Objective: Build the fastest car and race it on a track.<br>Lesson: Teaches aerodynamics, competition, and the iterative design process.<br>**The Earthquake-Proof Tower:**<br>Objective: Construct the tallest tower that can withstand a simulated earthquake.<br>Lesson: Focuses on stability, resilience, and innovation under constraints.<br>**The Survival Scenario:**<br>Objective: Build a structure or tool that would help survive a given scenario (stranded on an island, lost in space, etc.).<br>Lesson: Emphasizes out-of-the-box thinking and adaptability.<br>**The Color Code Communication:**<br>Objective: Teams build a structure where the colors of the bricks represent different elements of a business plan or strategy.<br>Lesson: Highlights non-verbal communication and the importance of clear messaging.<br>Each of these exercises not only uses the fun and familiar medium of LEGO bricks but also aligns with key business concepts like innovation, teamwork, resource management, and strategic planning. | |
| Pick one and draw an example finished outcome to inspire the team-building exercises | 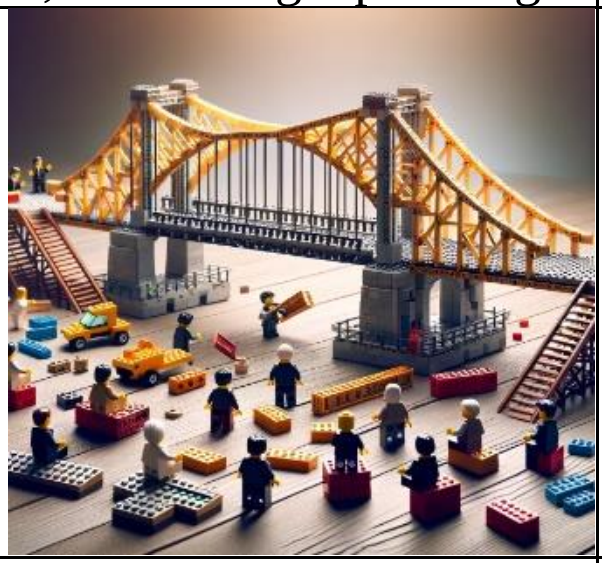 | MM Task: Image Generation and Self-Instruction |
| Modify the lesson plan to suit elderly patients with occupational therapy sessions about sorting and using their hands therapeutically | For occupational therapy sessions with elderly patients focusing on sorting and hand therapy using LEGO bricks, the lesson plan could be modified to be more therapeutic and accessible. Here's a tailored plan:<br><br>Objective: To improve fine motor skills, cognitive function, and hand-eye coordination through sorting and assembling LEGO bricks.<br><br>**Materials Needed:**<br>• A variety of LEGO bricks in different colors and sizes.<br>• Sorting trays.<br>• Tables with comfortable seating.<br>• Simple model examples for inspiration.<br>**Session Plan:**<br><br>*Warm-Up (5-10 minutes):* | MM Task: Modify Instructions Based on Audience |
| | • Begin with hand exercises to warm up the fingers and wrists. This could include stretching and gently massaging the hands.<br>*Color Sorting (10-15 minutes):*<br>• Spread a variety of LEGO bricks on the table.<br>• Ask patients to sort the bricks by color into provided trays.<br>• This exercise helps with cognitive engagement and fine motor control.<br>*Size Sorting (10-15 minutes):*<br>• After color sorting, proceed to sort the bricks by size.<br>• Emphasize gentle handling and precise movements to foster dexterity.<br>*Pattern Assembly (15-20 minutes):*<br>• Use simple patterns or images that require patients to select bricks by color and size to create a simple object (e.g., a flower or a house).<br>• Assist patients as needed to ensure they are not straining.<br>*Free Build (15-20 minutes):*<br>• Encourage patients to use the sorted bricks to create anything they wish.<br>• This encourages creativity and provides cognitive stimulation.<br>*Cool Down (5-10 minutes):*<br>• Conclude with a cool-down session, encouraging patients to share what they created with the group.<br>• Finish with hand stretches to relax the muscles after the activities.<br><br>**Key Considerations:**<br>• Ensure bricks are large enough to handle easily to prevent strain.<br>• Keep sessions lighthearted and stress-free; the focus is on therapy and enjoyment.<br>• Monitor patients for any signs of fatigue or frustration and adjust the activity accordingly.<br>• Encourage social interaction and storytelling about their creations to enhance the social aspect of the session.<br>By focusing on these areas, LEGO bricks can be an effective tool in occupational therapy for the elderly, promoting mental and physical well-being through engaging and rewarding activities. | |
| Create an IKEA inspired chair using Lego bricks | The image of an IKEA-inspired chair constructed from LEGO bricks has been created, featuring a modern and minimalist design. 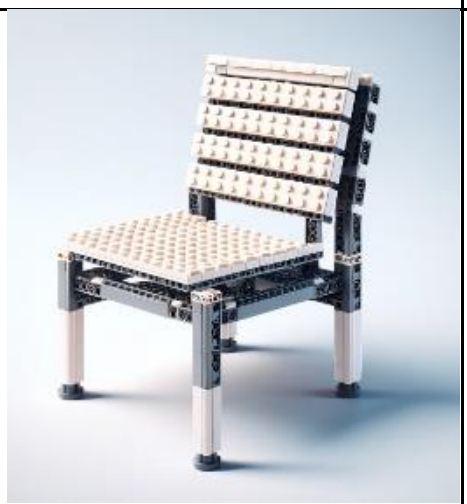 | MM Task: Image Style Transfer |
| Describe your design choices as if on an architectural jury and play the role of the student defending the design | As a student presenting this LEGO chair design, I would defend my choices by emphasizing the following aspects:<br>**Functionality and Aesthetics:**<br>The chair design embodies the principles of Scandinavian design that IKEA is known for—simplicity, minimalism, and functionality. It features clean lines and a neutral color palette, which makes it versatile for various settings. The chair is not just a model; it represents a functional piece of furniture that prioritizes user comfort and practicality.<br><br>**Material Efficiency:** | MM Task: Image to Text Description |
| | LEGO bricks are modular and reusable, making them an excellent choice for iterative design processes. This aligns with the sustainable practices in architecture, where materials can be repurposed or recycled. The design uses a minimal number of bricks to achieve maximum strength, reflecting an understanding of material properties and structural integrity.<br>**Innovative Use of LEGO:**<br>The use of LEGO in furniture design is innovative, demonstrating the versatility of the medium. LEGO bricks are typically associated with play, but here they are used to create something functional and stylish. This speaks to the potential of unconventional materials in design and their ability to surprise and inspire.<br>**Scale and Proportion:**<br>The design maintains proper scale and proportion, which are fundamental to architectural aesthetics. The slender legs and the flat plate construction for the seat and backrest reflect an understanding of human ergonomics, as well as the visual balance necessary in good design.<br>**Cultural Relevance:**<br>LEGO, a Danish invention, and IKEA, a Swedish company, share cultural roots in design philosophy. This chair represents a dialogue between these two iconic brands, reflecting the deep appreciation for Nordic design principles.<br>In summary, this LEGO chair is not only a tribute to the enduring design principles that IKEa embodies but also a statement on the adaptability and innovative potential of LEGO bricks beyond their traditional use. The chair is a harmonious blend of playfulness and practicality, celebrating the creativity and thoughtfulness that underpin great design. | |
| Now switch roles, become the design judge and ask 5 questions to the student to probe their understanding and evaluate their grade | As a judge on the architectural jury, I would pose the following questions to the student to probe their understanding and assess their design:<br><br>**Structural Integrity:** How have you ensured that your LEGO chair design maintains structural integrity, and could you explain the specific load-bearing elements?<br>**User Experience:** Considering ergonomics, how does your design accommodate the varied postures and comfort requirements of different users?<br>**Material Considerations:** LEGO bricks have their limitations and strengths. Could you discuss why you chose LEGO as your primary material and how it benefits your design functionally and aesthetically?<br>**Sustainability:** With sustainability being a critical factor in contemporary design, how does your use of LEGO bricks align with principles of sustainable design?<br>**Innovation and Originality:** What aspects of your design do you consider innovative, and how does it push the boundaries of traditional furniture design?<br>The responses to these questions would help evaluate the student's depth of knowledge in design principles, problem-solving skills, and their ability to think critically about their work. | MM Task: Image to Question Series |
| Is there a real world building material and principles to explore legos in actual modular | Yes, the principles of LEGO can indeed be translated to real-world modular construction, particularly in the context of multifunctional urban furniture like a park bench that serves the homeless. | MM Task: Image to Abstract |
| construction of say a park bench for homeless that changes function from night to day | **Real-World Building Material:**<br>In the real world, modular construction often uses prefabricated units that can be easily assembled and disassembled. Materials such as reinforced polymers, precast concrete, and modular metal components would be suitable for outdoor furniture due to their durability, weather resistance, and ease of maintenance.<br>**Design Principles**:<br>The key principles from LEGO that apply to modular construction include:<br>• **Interlocking Mechanisms:** Much like LEGO bricks, modular components can be designed to interlock, ensuring easy assembly and stability without the need for additional fastening materials.<br>• **Customization:** Modular units can be rearranged or replaced as needed, allowing for customization of the bench's form and function.<br>• **Scalability:** Modular design allows for the bench to be extended or contracted, similar to how LEGO structures are built, accommodating varying numbers of users.<br>• **Functionality for Homeless Support:** A bench designed for homeless individuals might incorporate features such as:<br>  • *Convertibility:* The bench could convert into a temporary shelter at night, with fold-out elements that provide cover and protection.<br>  • *Storage:* Hidden compartments could store personal belongings securely, or contain blankets and necessities for cold nights.<br>  • *Ergonomics:* The design would need to consider comfort for sitting and lying down, possibly with adjustable surfaces.<br>• **Sustainable Design:** Modular construction facilitates sustainability through:<br>  • *Reusable Materials:* The bench components could be designed for reuse or repurposing, minimizing waste.<br>  • *Local Fabrication:* Modular parts can be manufactured locally, reducing the carbon footprint associated with transportation.<br>  • *Maintenance and Repair:* Individual damaged modules can be replaced without discarding the entire bench, conserving resources.<br>In summary, the LEGO-inspired principles of modularity, interactivity, and adaptability can be effectively applied to create multifunctional urban furniture that serves the dual purpose of public seating and nighttime shelter for the homeless. This would not only address urban design challenges but also provide a tangible solution to social issues. | Design Principles |
| Design a data center in the theme of a Rubik's cube (Midjourney) | 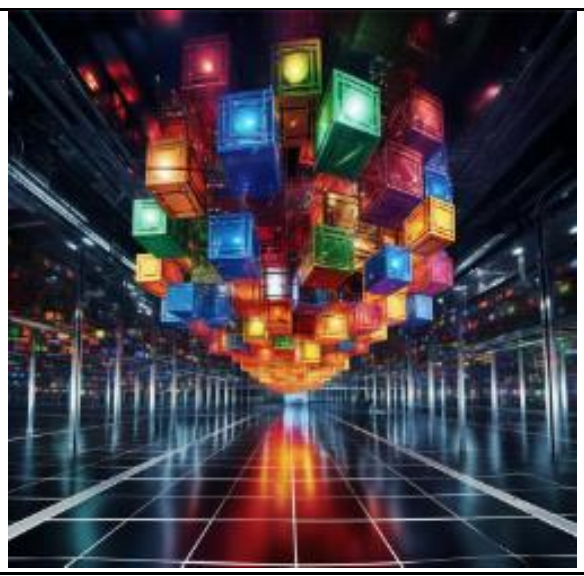 | MM Task: Image Style Transfer |
| Design a data center in the theme of a Rubik's cube (Midjourney) but substituting "light mode" theme for "dark mode" | 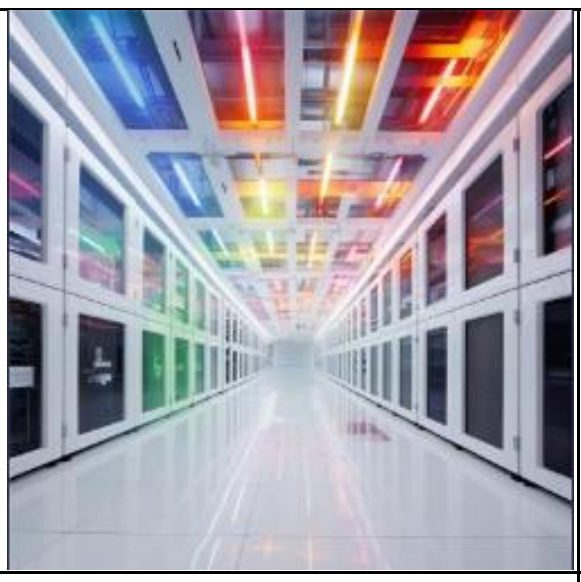 | MM Task: Image Style Transfer |
| Design a data center in the theme of a Rubik's cube (Midjourney) but change the color theme to satisfy color-blind restrictions | 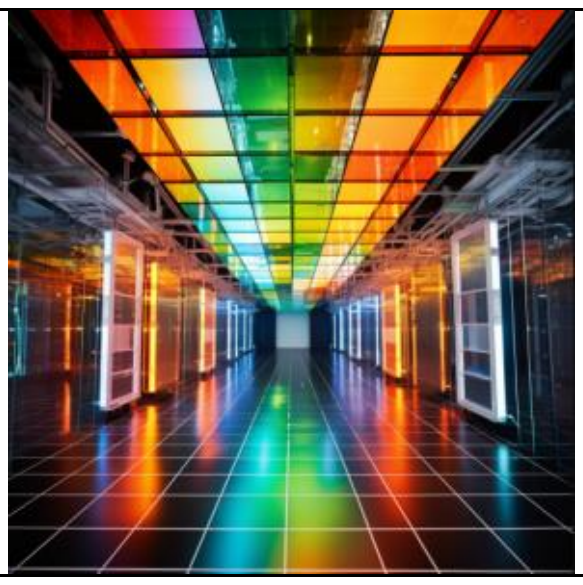 | MM Task: Image Style Transfer |

## APPENDIX B: AI AEROSPACE DESIGNER

| PROMPT | RESPONSE | NOTES |
|---|---|---|
| Let's make a paper airplane design for aerial tricks | 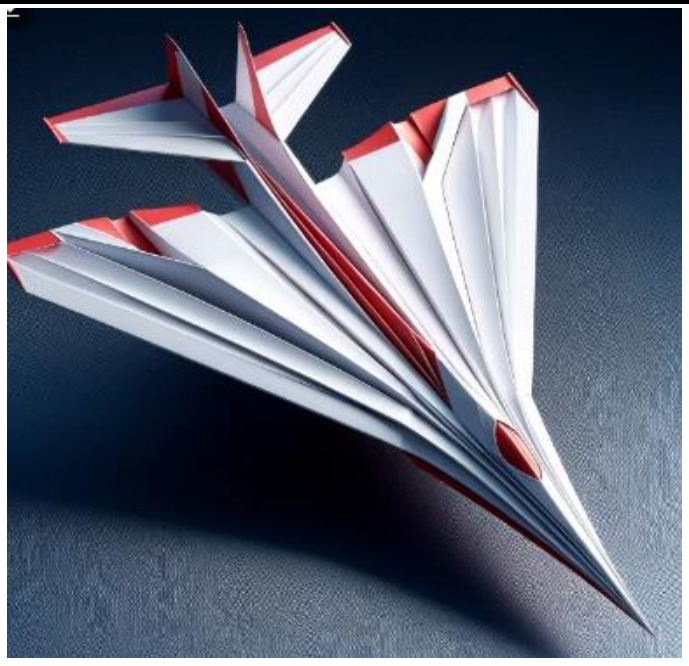 Here's a design for a highly aerodynamic paper airplane, tailored for performing aerial tricks. It features a sleek, streamlined design with reinforced wings and an innovative tail for precise control and maneuverability. The bright red and white color scheme makes it visually striking. This design is ideal for advanced paper airplane enthusiasts looking to enhance their trick performance. | MM Task: Image Generation |
| Give folding instructions to make it | Creating a paper airplane that closely resembles the designed model may be challenging, as the design includes elements like reinforced wings and a specialized tail, which are difficult to replicate exactly with paper folding. However, I can guide you through making a paper airplane that incorporates some aspects of the design, such as a sleek body and stable wings for tricks.<br>**Materials Needed:**<br>• A sheet of paper (preferably A4 or letter size)<br>**Folding Instructions:**<br>1. **Start with a Rectangular Sheet:** Place the paper on a flat surface with the longer sides at the top and bottom.<br>2. **Fold in Half Lengthwise:** Fold the paper in half lengthwise, then unfold it to have a central crease.<br>3. **Create the Nose:** Fold the top two corners down to the center crease, forming a triangle shape at the top.<br>4. **Form the Body:** Fold the angled edges of the triangle you just made into the center line, creating a slender, pointed nose.<br>5. **Reinforce the Nose:** Fold the pointed tip up about 2 cm, then fold it back down, aligning it with the previous folds. This creates a reinforced nose, adding weight and strength. | MM Task: Image to Text Instructions |
| | 6. **Fold the Plane in Half:** Fold the airplane in half along the central crease, with all the folds on the outside.<br>7. **Create Wings:** Fold each wing down so that the edge of the wing lines up with the bottom of the airplane's body. The wings should be slightly angled upward for stability.<br>8. **Add Wing Folds:** For additional stability, make small folds at the back of each wing, pointing upwards. These act like ailerons and help in maneuvering.<br>9. **Adjust and Fine-Tune:** After the basic shape is ready, you might need to make slight adjustments to the wings or nose to ensure the plane flies well. The plane should have a slight upward angle on the wings when held flat.<br>Remember, the performance of your paper airplane can be influenced by the type of paper, the precision of your folds, and the symmetry of the design. Happy flying! | |
| Show all the needed folds on a flat A4 paper rendering | 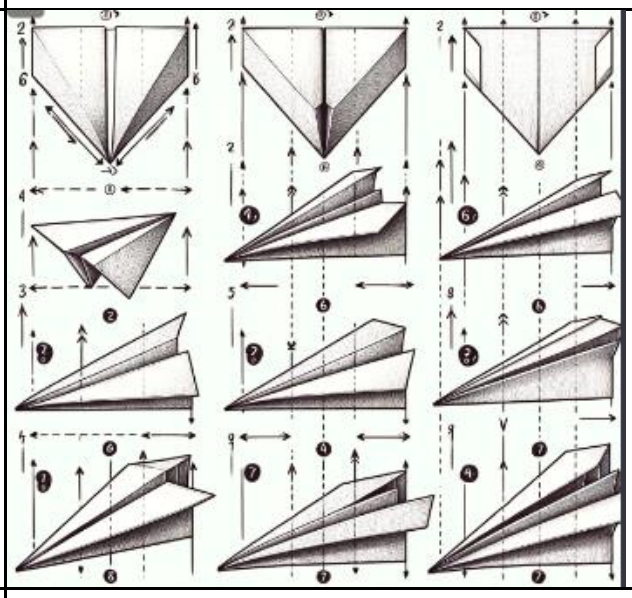 Here's a step-by-step guide illustrating the folds needed on a flat A4 paper to create an aerodynamic paper airplane. The image shows each fold clearly with labels and arrows indicating the direction of each fold. This should help you in accurately folding your paper airplane. | MM Task: Text to Image Instruction |
| Make 3 images in sequence to illustrate key building stages | Here are three images, each illustrating a key stage in the folding process of making an aerodynamic paper airplane from a flat A4 paper. These images show the progression from one step to the next, highlighting the important folds and shapes at each stage. 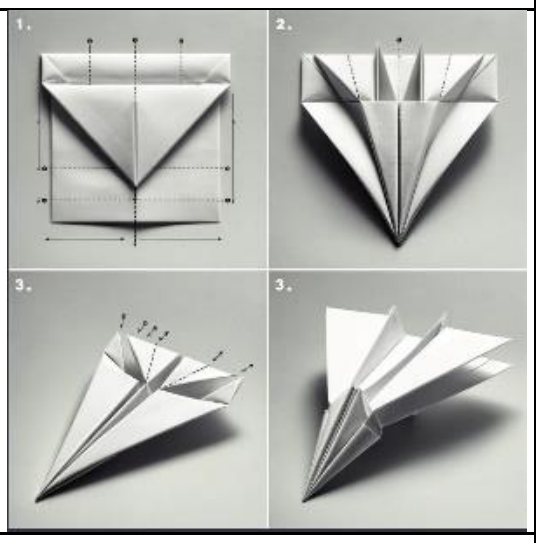 | MM Task: Instructions to Temporal Movie |
| Make that 2x2 image grid into a gif by cropping each corner and stacking them in order to build a gif | Movie available online<br>https://deeperbrain.com/challenge/paper_airplane_folding.GIF | MM Task: Complete Movie Instruction |

## APPENDIX C: IMAGE TO INGREDIENT AI BARTENDER

| PROMPT | RESPONSE | NOTES |
|---|---|---|
| 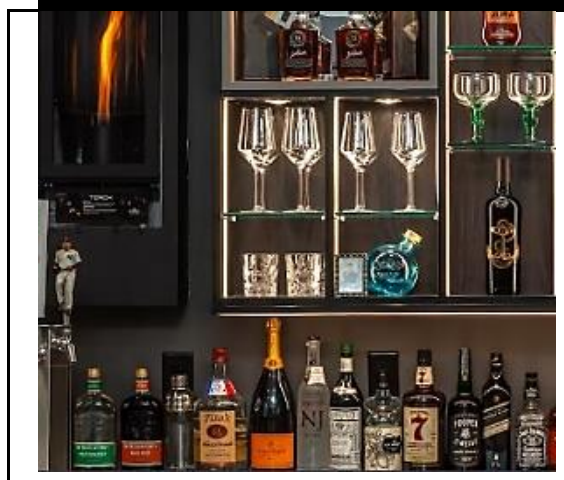 based on the image of this home bar, give me some innovative mixes from the identified | The image of your home bar displays a selection of spirits that can be used to craft a variety of innovative cocktails. Here are some ideas for mixes and their pairings based on the visible bottles:<br>1. **Whiskey Sour Twist**:<br>• Mix whiskey with fresh lemon juice, a dash of simple syrup, and a splash of orange juice for a citrusy twist on the classic Whiskey Sour. Garnish with an orange wheel and a cherry.<br>• Pairing: This cocktail goes well with savory appetizers like smoked salmon canapés or bacon-wrapped dates.<br>2. **Spiced Rum Old Fashioned**: | MM Task: Image to Ingredients to Novel Constructions |
| bottles and their best pairings for innovative cocktails. Assume I store various fruit juices in a refrigerator not shown | • Use spiced rum as the base, add a sugar cube, a couple of dashes of Angostura bitters, and a splash of water. Stir until the sugar dissolves, and garnish with a twist of orange peel.<br>• Pairing: The spiciness of the rum complements sweet desserts like chocolate fondue or caramel-drizzled cheesecake.<br>3. **Gin and Tonic with a Twist**:<br>• Combine gin with tonic water and add a splash of grapefruit juice for a tart edge. Garnish with a sprig of rosemary and a slice of grapefruit.<br>• Pairing: This refreshing drink is perfect with light fare such as a shrimp cocktail or a citrus-arugula salad.<br>4. **Tequila Sunrise Remix**:<br>• Blend tequila with a mix of fresh orange juice and a hint of pineapple juice. Layer in a splash of grenadine for the sunrise effect and garnish with a pineapple wedge.<br>• Pairing: Enjoy this tropical cocktail with grilled fish tacos or a zesty mango salsa with tortilla chips.<br>5. **Vodka Infusion**:<br>• Infuse vodka with fresh berries (if available) for a few hours. Serve chilled with a dash of lime juice and a splash of cranberry juice, topped with sparkling water.<br>• Pairing: The berry-infused vodka pairs delightfully with chocolate truffles or a fruit tart.<br>6. **Bourbon Peach Smash**:<br>• Muddle fresh peach slices in a glass, add bourbon, a bit of honey syrup, and a dash of lemon juice. Top with crushed ice and a sprig of mint.<br>• Pairing: This Southern-inspired cocktail is excellent with barbecue dishes or a peach cobbler dessert.<br>Remember to adjust the proportions to your taste and the size of your glasses. For a more personalized touch, consider using herbs from a garden or unique bitters to add depth to your cocktails. Enjoy experimenting with these suggestions! | |
| On what shelf do you see the Vodka for recipe 5? | The vodka appears to be located on the bottom shelf of your home bar, based on the image provided. It's typically stored alongside other base spirits like gin, rum, and whiskey, which are also visible in your selection. For the recipe I suggested, you can use any vodka you have on hand, whether it's plain or flavored, depending on your preference for the infusion. | MM Task: Object Detection |
| What to do with the Jack Daniels? | Jack Daniel's, being a Tennessee whiskey, has a distinct flavor that can be enjoyed neat, on the rocks, or in a variety of mixed drinks. Here are a few cocktail ideas:<br>1. **Classic Jack & Coke**:<br>• Simply mix Jack Daniel's with cola over ice and garnish with a slice of lime for a refreshing classic.<br>2. **Tennessee Honey Tea**:<br>• If you have Jack Daniel's Tennessee Honey, mix it with iced tea. Add a squeeze of lemon for a sweet and tangy beverage.<br>3. **Lynchburg Lemonade**:<br>• Combine Jack Daniel's with triple sec, lemon juice, and a splash of soda. This drink is both tangy and refreshing, perfect for a hot day.<br>4. **Jack Daniel's Manhattan**: | MM Task: Novel Text Reconstruction |
| | • Mix Jack Daniel's with sweet vermouth and a few dashes of bitters. Garnish with a cherry for a Southern twist on the traditional Manhattan.<br>5. **Whiskey Sour**:<br>• Shake Jack Daniel's with lemon juice and simple syrup. Serve over ice and garnish with a cherry and an orange slice.<br>6. **Apple Jack**:<br>• Mix Jack Daniel's with apple juice or apple cider. Garnish with a slice of apple and a cinnamon stick for a fall-inspired drink. | |

## APPENDIX D: IMAGE TO INGREDIENT AI CHEF

| PROMPT | RESPONSE | NOTES |
|---|---|---|
| 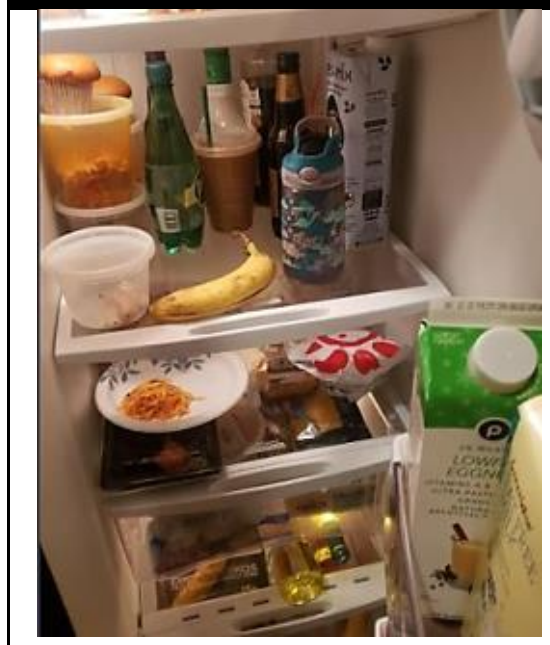<br>Here' s a picture of my refrigerator. analyze the contents and make a recipe for a packed lunch that includes only the ingredients seen | Based on the contents of the refrigerator, here's a recipe for a packed lunch:<br>**Chilled Pasta Salad with Tuna**<br>**Ingredients:**<br>• Cooked pasta (looks like some is already prepared in a container)<br>• Canned tuna (assuming there's a can hidden somewhere in the fridge or pantry)<br>• A ripe banana (for a sweet side or dessert)<br>• Olive oil (bottle in the door)<br>• Any available fresh or pickled vegetables (there appear to be some vegetables in the lower drawer)<br>• Salt and pepper for seasoning (staple ingredients not shown but commonly available)<br>**Instructions:**<br>1. Take the pre-cooked pasta out of the fridge and let it come to room temperature.<br>2. If you have fresh vegetables like tomatoes or cucumbers, chop them into bite-sized pieces. If you have pickles or any other pickled vegetables, dice them as well.<br>3. Drain the canned tuna and mix with the pasta in a large bowl.<br>4. Add the chopped vegetables to the pasta and tuna.<br>5. Drizzle olive oil over the pasta mixture. Toss everything to coat evenly. Season with salt and pepper to taste.<br>6. Peel the banana and place it whole or sliced alongside your pasta salad for a touch of sweetness.<br>7. Pack the salad in a lunch container and enjoy chilled or at room temperature.<br>Note: If there are any dressings or condiments, you could also add those to your pasta salad for extra flavor. Additionally, if there's cheese in the fridge, you could shred some and sprinkle it over the top for added richness. | MM Task: Image to Object Detection to Text Generation |
| revisit the recipe since the tuna will not keep well in a packed lunch during summer | Understood, let's revise the recipe for a packed lunch that will hold up well in the heat.<br>**Mediterranean Chickpea Salad**<br>Assuming you might have a can of chickpeas in your pantry to complement what's in your fridge, here's a recipe:<br>Ingredients:<br>• 1 can of chickpeas, drained and rinsed (substitute for tuna) | MM Task: Text Answer Refinement under Complex Instructions |
| | <ul><li>Cooked pasta</li><li>Olive oil (bottle seen in the door)</li><li>Fresh or pickled vegetables (seen in the drawers, like cucumbers or peppers)</li><li>Salt and pepper for seasoning</li><li>Any available herbs (if you have any fresh herbs or dried ones in your pantry)</li><li>Lemon juice (if you have a lemon or bottled lemon juice in the fridge)</li><li>A ripe banana for dessert</li></ul>**Instructions:**<ol><li>In a large bowl, combine the cooked pasta and chickpeas.</li><li>Chop any fresh vegetables you have into bite-sized pieces and add them to the bowl.</li><li>If using pickled vegetables, dice them and include them as well.</li><li>Drizzle olive oil over the mixture, add a squeeze of lemon juice for some acidity, and season with salt, pepper, and herbs to your liking.</li><li>Toss everything together until it's well mixed.</li><li>The banana can be kept whole or sliced right before eating to prevent browning.</li><li>Pack the salad in a lunch container. This dish can be served cold or at room temperature, making it suitable for a packed lunch in the summer.</li></ol>This Mediterranean Chickpea Salad is hearty, refreshing, and less likely to spoil when compared to a salad with tuna. It's packed with protein from the chickpeas and will hold up well even on a hot day. Enjoy your packed lunch! | |
| remove the highest calorie ingredient and reformulate the packed lunch ideas | The highest-calorie ingredient in the proposed recipe is likely to be the olive oil, especially if used generously. To create a lower-calorie packed lunch that will keep well, we can omit the olive oil and focus on the other ingredients available.<br>**Low-Calorie Chickpea Pasta Salad**<br>Ingredients:<ul><li>1 can of chickpeas, drained and rinsed</li><li>Cooked pasta (use a moderate amount to control calories)</li><li>Fresh or pickled vegetables (such as cucumbers, peppers, or any others available in the lower drawer)</li><li>Lemon juice (if available, for dressing)</li><li>Vinegar (if you have any in your pantry, as a substitute for olive oil)</li><li>Salt and pepper for seasoning</li><li>Herbs (any fresh or dried ones you might have for added flavor)</li><li>A ripe banana for dessert</li></ul>**Instructions:**<ol><li>In a mixing bowl, combine the drained chickpeas with the cooked pasta.</li><li>Chop your fresh vegetables into bite-sized pieces and add them to the bowl.</li><li>If you have pickled vegetables, add those as well for extra flavor without additional calories.</li></ol> | MM Task: Text Answer Refinement under Complex Instructions |
| | 4. Instead of olive oil, use a small amount of vinegar mixed with lemon juice to dress the salad. This will add flavor without the fat content.<br>5. Season the salad with salt, pepper, and herbs to your taste.<br>6. Toss the salad well to ensure that the dressing and seasonings are evenly distributed.<br>7. The banana will serve as a naturally sweet dessert with no added sugars.<br>8. Pack the salad in your lunch container; it’s ready to be enjoyed cold or at room temperature, perfect for a summer lunch that needs to last outside of the fridge.<br>By removing the olive oil and controlling the portion of pasta, you significantly reduce the calorie content of your packed lunch while still having a satisfying and refreshing meal. | |
| reimagine lunch without utensils | Certainly! If you need a lunch that doesn't require utensils, finger foods are the way to go. Here's a suggestion based on the contents of your fridge:<br>**Banana and Vegetable Wraps with Chickpea Dip**<br>**Ingredients:**<br>• Tortillas or any wrap-friendly flatbread (if not available, lettuce leaves can be a good substitute)<br>• Chickpeas, mashed into a dip<br>• Fresh vegetables (such as cucumber or peppers), sliced into sticks for easy wrapping<br>• Salt, pepper, and any available herbs or spices for flavoring the chickpea dip<br>• A ripe banana, for a sweet treat<br>**Instructions:**<br>1. Mash the chickpeas in a bowl using a fork or a potato masher. Season with salt, pepper, and herbs to taste. If you have lemon juice, a splash can add a nice zest to your dip.<br>2. Slice the fresh vegetables into sticks or strips that will be easy to handle.<br>3. Lay out a tortilla or a large lettuce leaf. Spread a spoonful of the chickpea mash onto the tortilla or lettuce.<br>4. Place a few vegetable sticks on top of the chickpea spread.<br>5. Roll the tortilla or lettuce leaf around the filling to create a wrap that can be eaten with your hands.<br>6. The banana can be left whole or cut into chunks, depending on what you find easier to eat without utensils.<br>7. Pack the wraps and banana separately in your lunch container.<br>With this meal, you've got a portable, nutritious lunch that doesn't require any utensils. The chickpea dip provides a creamy, flavorful base that complements the crunch of fresh vegetables, and the banana offers a quick, energizing dessert. Enjoy your utensil-free lunch! | MM Task: Text Answer Refinement under Complex Instructions |
| i need you to be strict about only the ingredients seen in the picture. Do not imagine ingredients not seen and continue to follow the constraints previously | Acknowledged. Using only the visible ingredients from your refrigerator and maintaining the no-utensils and summer-friendly criteria, here's a revised packed lunch idea:<br>**Chilled Pasta and Banana Finger Food**<br>**Ingredients:**<br>• Pre-cooked pasta (from the container)<br>• Fresh vegetables (looks like there might be some in the bottom drawer, such as cucumber or bell pepper) | MM Task: Text Answer Refinement under Complex Instructions |
| on this packed summer lunch | • A ripe banana<br>**Instructions:**<br>1. If the vegetables in the bottom drawer are indeed cucumber or bell pepper, slice them into long strips that can be picked up easily with fingers.<br>2. Take the pre-cooked pasta from the container. If it's in a shape that can be eaten without utensils, such as penne or fusilli, it's perfect. If it's a longer pasta like spaghetti, consider cutting it into shorter lengths with scissors or by hand.<br>3. Mix the pasta and vegetable strips in a container. If there's any condiment like a vinaigrette or lemon juice in the fridge that's not visible in the picture, you could toss the pasta and vegetables in it. If not, they can be enjoyed plain or with any available seasoning like salt and pepper.<br>4. Place the ripe banana on the side. When it's time to eat, the banana can be peeled and eaten as is, serving as a refreshing and sweet end to the meal.<br>**Packing Tips:**<br>• Keep the banana whole to avoid browning and maintain freshness.<br>• Place the pasta and vegetable mix in a container that can be shaken to distribute any hidden seasoning or minimal residual olive oil from the pasta.<br>• You could use any sturdy food like the bell pepper strips as an edible 'utensil' to scoop up the pasta.<br>This no-cook, utensil-free lunch idea respects the constraints and uses only the visible items in the refrigerator picture. It's suitable for summer as there are no ingredients that spoil quickly, and it's all easy to eat with your hands. Enjoy your simple, on-the-go summer lunch! | |
| How is pasta possible with no utensils? | You're right; pasta typically requires utensils. However, if we are to use the pasta in a manner that doesn't require utensils, we could consider making pasta bites that are easy to pick up. Assuming the pasta in the container is already cut into pieces or uses shapes that are bite-sized, such as penne, farfalle, or fusilli, here's an idea:<br>**Pasta Snack Bites**<br>**Ingredients:**<br>• Pre-cooked pasta (from the container, assuming it's bite-sized)<br>• Any condiment or sauce seen in the fridge door (if applicable for flavoring)<br>**Instructions:**<br>1. If the pasta is not already in bite-sized pieces, you could try to break or cut the pasta into smaller pieces that are easy to pick up. If the pasta is something like spaghetti or linguine, this might not be feasible without it becoming too messy.<br>2. If there's a thicker sauce or dressing visible in the fridge (perhaps in the jars on the door), you could lightly toss the pasta in it to add flavor. This would only work if the sauce can stick to the pasta and is not too liquidy, as that would require a utensil to eat without making a mess.<br>3. After flavoring, you can place the pasta bites into a container. At lunchtime, the pasta can be eaten with fingers like a snack.<br>**For the vegetables:**<br>• Slice any solid vegetables like cucumbers or bell peppers into sticks or chunks that can be eaten with fingers. | MM Task: Text Answer Refinement under Complex Instructions |
| | **For the banana:**<br>• Keep it whole and peel it when ready to eat as a sweet finish to your meal.<br>In this way, you can enjoy a simple, no-utensil-required lunch with the available ingredients. This suggestion works under the assumption that the pasta is in a form that is manageable without utensils, and there's a suitable condiment to add flavor without needing additional dressing. If the pasta is not in a suitable form, then we may have to reconsider the options based on the available ingredients. | |
| I don't want you to imagine anything not seen. I also don't think cooked pasta can be broken into bite size finger food. | Understood. We'll work strictly with what's visible in the fridge and the constraints given.<br>**Banana & Vegetable Snack Box**<br>**Ingredients:**<br>• A ripe banana<br>• Vegetables (from the bottom drawer; if there are items like carrot sticks, cucumber, or small tomatoes, these would work well)<br>**Instructions:**<br>1. Carefully remove the vegetables from the bottom drawer. If they are in a form that is naturally bite-sized or can be easily picked up, like baby carrots, small tomatoes, or snap peas, they are ready to go. If they are larger vegetables, they may need to be sliced, but since utensils are not preferred for this lunch, we will set aside any vegetables that can't be easily hand-held.<br>2. Place the selected vegetables and the whole banana into a lunch box. They can be arranged in separate compartments or together, depending on preference.<br>Since we're avoiding the use of utensils and the pre-cooked pasta may not be practical for a finger food lunch, the packed lunch will consist solely of the banana and any snackable vegetables from your fridge, which can be eaten as they are. This keeps your lunch simple, utensil-free, and suitable for summer weather. Enjoy your fresh and straightforward lunch! | MM Task: Text Answer Refinement under Complex Instructions |
| Let's revisit the packed lunch with a new idea that nothing will be eaten by the owner of the lunch, but that will instead bring maximum trading value with other food provided at the table in a barter system | Considering the barter potential of your packed lunch items, it's important to focus on items that others might find appealing and might be willing to trade with. Here's a new approach:<br><br>**High-Value Barter Lunch Pack**<br>**Ingredients:**<br>• Pre-cooked pasta (assuming it's a popular dish like a pasta salad that's seasoned and ready to eat)<br>• A ripe banana (bananas are a popular and healthy sweet snack)<br>• Beverages (if those bottles contain artisanal drinks, they could be quite valuable for trading)<br>• Any whole fruits or packaged items (not visible but typically have good trade value)<br>**Instructions:**<br>1. Keep the pre-cooked pasta in its container. If it's a desirable dish like a well-made pasta salad, it could have significant trade value.<br>2. Include the ripe banana; as a piece of whole fruit, it's a desirable item for someone looking for a healthy option or a sweet treat.<br>3. If those bottles are specialty beverages, they could be very attractive for trade. Specialty drinks often have a higher perceived value, especially if they're from a sought-after brand or flavor. | MM Task: Text Answer Refinement under Complex Instructions |
| | 4. Any other whole fruits, packaged snacks, or treats that you might have that are not visible in the fridge could be added to increase the variety of your barter pack.<br>**Packing Tips:**<br>• Pack the items in such a way that they are visible and look appealing. Presentation can affect the perceived value of the items.<br>• Keep the banana at the top to avoid bruising.<br>• If the pre-cooked pasta is in a see-through container, that would be ideal. If not, consider transferring it to one if possible.<br>When trading, start with items that you value the least or that you have in abundance. Keep the more valuable items, like specialty beverages or the banana, for trades that are worth it. The goal is to maximize the return on what you're offering, so assess what others have and what they seem to value the most before making an offer. Happy bartering! | |
| generate an example image of the unpacked lunch we finally concluded with | 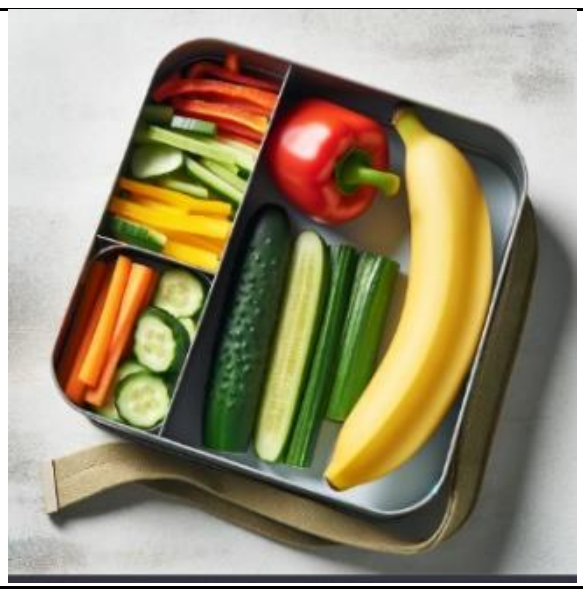 Here is an example image of the unpacked lunch we discussed, featuring a whole ripe banana and an assortment of raw vegetables. | MM Task: Task: Image Generation |

## APPENDIX E: IMAGE TO AI ORIGAMI (JAPANESE ART OF FOLDING)

| PROMPT | RESPONSE | NOTES |
|---|---|---|
| 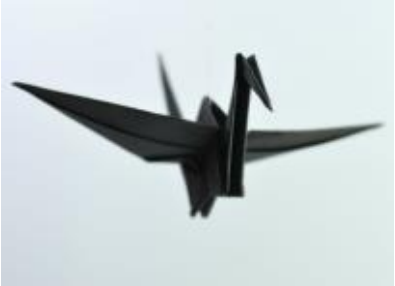<br>What is this image? | The image appears to be of an origami crane. Origami is the Japanese art of paper folding, and the crane is one of its most iconic creations. It's often associated with peace and good fortune. This particular crane seems to be made from black paper and is suspended, possibly for decorative purposes. | MM Task: Image to Object Detection |
| give an instructional manual on how best to generate the crane origami figure | Creating an origami crane involves a series of folds, which can be done on almost any square piece of paper. Here's a simplified set of instructions to guide you through the process:<br>**Materials Needed:**<br>• One square piece of paper - traditionally origami paper is used, but any paper will do as long as it's not too thick.<br>**Instructions:**<br>1. **Preliminary Fold:**<br>• Place the paper square in front of you so that it looks like a diamond.<br>• Fold the paper in half diagonally to form a triangle, then unfold.<br>• Fold it diagonally in the opposite direction and unfold, so you have an X crease.<br>2. **Square Base:**<br>• Flip the paper to the other side.<br>• Fold the paper in half, top to bottom, then unfold.<br>• Fold in half left to right, then unfold.<br>• Using the creases, bring the top three corners down to the bottom corner, collapsing the paper into a smaller square.<br>3. **Bird Base:** | MM Task: Text Answer Refinement under Complex Instructions |
| | • Lift one layer from the bottom corner up to the top, creasing at the base so that you form a triangle.<br>• Flip the paper over and repeat this step on the other side, forming a diamond shape.<br>• Fold the left and right corners of the upper layer to the top corner, creating a kite shape.<br>• Flip the paper over and repeat on the other side.<br>4. **Creating Wings and Neck:**<br>• Lift the bottom of the top flap and fold it upwards, reversing the crease on the side of the diamond to form a narrow kite shape.<br>• Do this on both sides, ensuring that the open part of the paper is facing away from you.<br>5. **Shaping the Head:**<br>• Fold down one of the narrow kite flaps (this will become the crane's head) and crease.<br>• Pull the beak out to give it a three-dimensional shape.<br>6. **Finishing the Wings:**<br>• Pull the two long sides apart from the bottom, flattening out what will become the wings.<br>• Gently pull the wings to crease the folds firmly.<br>7. **Final Touches:**<br>• Carefully pull the wings to make sure they're even.<br>• Optionally, you can puff out the body by gently blowing into the bottom part of the crane.<br>Remember, origami takes practice, so don't be discouraged if your first few attempts aren't perfect. Happy folding! | |
| can you convert it to a peacock in an origami image? | 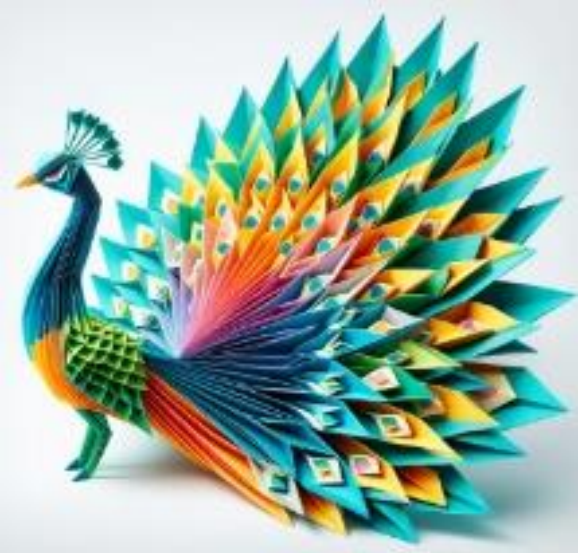 Here is the generated image of an origami peacock with its tail fanned out | MM Task: Image Generation |
| That's complex. Give a minimum fold to turn the origami crane into a peacock in the simplest way | Transforming an origami crane into a simplified version of a peacock can be done by modifying the tail to resemble the fanned tail feathers of a peacock. Here's a simple method to do this with minimal additional folds:<br>1. **Start with your Origami Crane**: Have your completed origami crane ready.<br>2. **Modify the Tail**:<br>• Carefully open up the tail end of the crane to make it possible to fan out.<br>• Crease each side of the tail diagonally to make small, fan-like folds. These will not be traditional origami folds but more like crimps.<br>• Spread out the crimps to simulate the fanned tail of a peacock.<br>3. **Adjust the Wings**:<br>• Bring the wings slightly downwards to align with the new tail shape and give it a more peacock-like silhouette.<br>4. **Shape the Head**:<br>• If your crane has a pointed head, you can fold a small portion of the tip inside to create the rounded head shape typical of a peacock. | MM Task: Instruction Refinement |
| | • Add a small crease or bend at the top of the head to represent the peacock's crest.<br>Remember, this will be a very abstract version of a peacock, as real origami peacocks can be quite intricate and may require starting from scratch with a different folding sequence. But these modifications can give your crane a hint of a peacock's character. | |
| draw the version you gave instructions to build of the minimal peacock derived from the crane | 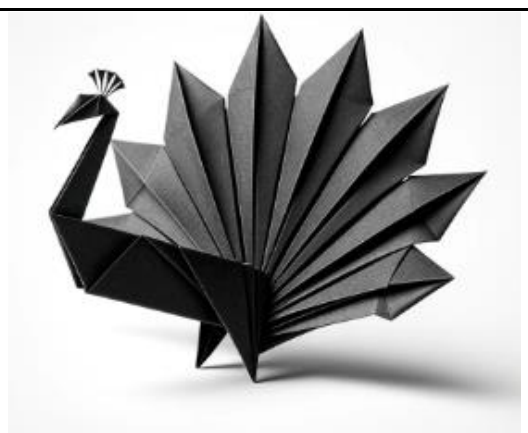 Here is the generated image of a simple origami peacock, modified from an origami crane with a fanned tail and adjusted features to suggest a peacock's silhouette. | MM Task: Image from Instructions |

## APPENDIX F: IMAGE TO AI KINTSUGI (JAPANESE ART OF REPAIRING BROKEN OBJECTS)

| PROMPT | RESPONSE | NOTES |
|---|---|---|
| Act as a hobbyist and plastic model car builder. Give me a futuristic design for a flying car with attention to also supporting building instructions and kit manufacturing. | **Model Name:** SkyBlade X1<br>**Design Inspiration:** The SkyBlade X1 draws inspiration from sleek, aerodynamic sports cars and sci-fi aesthetics. Its design combines the elegance of a luxury automobile with the futuristic elements of a flying vehicle, creating a stunning and unique model for hobbyists.<br>**Materials:** To make the model more accessible and affordable for hobbyists, we'll use lightweight plastic components for the body and wings. The internal structure will consist of a combination of ABS and polycarbonate parts for strength and durability.<br>**Scale:** 1:24 scale is a popular choice for model cars, and it allows for a reasonable level of detail while keeping the size manageable.<br>**Key Features:**<br>1. **Sleek Exterior:** The SkyBlade X1 features a low-profile, streamlined body with smooth curves and sharp lines for an ultra-modern look.<br>2. **Retractable Wings:** The model includes retractable wings that can be extended or retracted to simulate the transformation from car to aircraft mode.<br>3. **Scissor-Style Doors:** Just like many sports cars, the SkyBlade X1 has iconic scissor-style doors that open upward for added visual appeal.<br>4. **Transparent Cockpit Canopy:** The canopy is made from transparent plastic, allowing a clear view of the detailed interior, including seats, control panels, and a miniature pilot figure.<br>5. **LED Lights:** LED lights are integrated into the front and rear of the model, creating a realistic lighting effect that can be controlled by a switch.<br>6. **Adjustable Landing Gear:** The landing gear can be adjusted to simulate takeoff and landing, providing an engaging interactive element.<br>**Building Instructions:**<br>1. **Gather Materials:** Start by collecting all the necessary materials, including plastic sheets, rods, glue, and paint in the chosen color scheme. | MM Task: Text to Instructions |

| | | | |
|---|---|---|---|
| | 2. **Assemble Chassis:** Begin by assembling the chassis and interior components, including the seats and control panels.<br>3. **Body Construction:** Next, build the body of the car, paying attention to the sleek curves and lines. Attach the scissor doors and ensure they open and close smoothly.<br>4. **Wings and Landing Gear:** Construct the retractable wings and adjustable landing gear, ensuring they fit snugly onto the chassis.<br>5. **Canopy and LED Installation:** Attach the transparent canopy and integrate the LED lights into the front and rear sections. Wire them to a control switch for easy operation.<br>6. **Finishing Touches:** Paint the model using your chosen color scheme, adding any decals or logos for a personalized touch. Ensure that the pilot figure fits comfortably in the cockpit.<br>7. **Test and Display:** Test all moving parts, such as the retractable wings and scissor doors, to ensure they function correctly. Once satisfied, display your SkyBlade X1 in your collection or diorama.<br>**Kit Manufacturing Considerations:** When manufacturing kits for the SkyBlade X1, consider packaging the components in a well-organized manner, with clear step-by-step instructions and labeled parts. Include pre-cut plastic sheets, connectors, screws, and any other necessary hardware. Additionally, provide different color options for the model's body to cater to different preferences.<br>With attention to detail, user-friendly instructions, and quality materials, the SkyBlade X1 model kit will be an exciting and accessible project for hobbyists who love both cars and aviation. | | |
| Draw a futuristic flying car according to these instructions | Here is the image of the futuristic flying car model named SkyBlade X1. 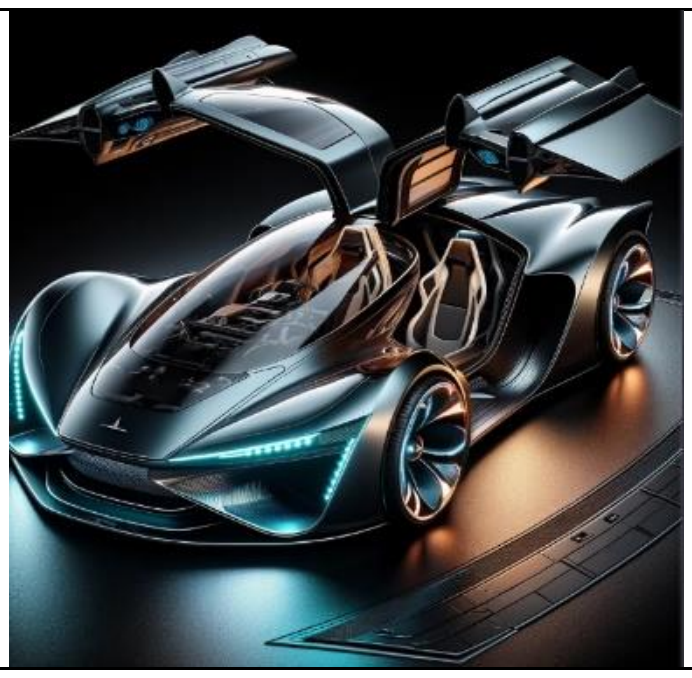 | MM Task: Text to Image Generation | |
| Let's practice *Kintsugi* (Japanese art of repairing broken objects) on this model. Assume you get the kit and are required to make something entirely different from its parts, not following the original instruction kit. Draw that imagined object from the parts list but with the requirement that it cannot be a car. | 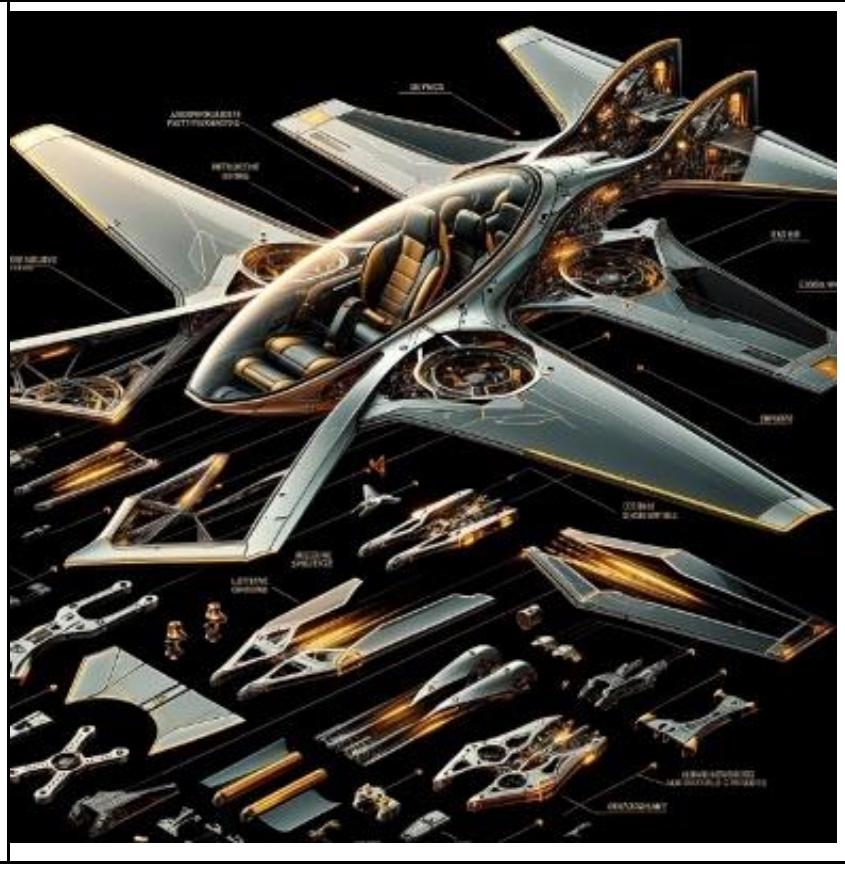 Here is the imagined object created from the parts of the SkyBlade X1 model kit | MM Task: Text to Image Refinement from Parts List | |

| | | |
|---|---|---|
| From those two images, identify one part using a red mask that is common to both the original and the reimagined design | 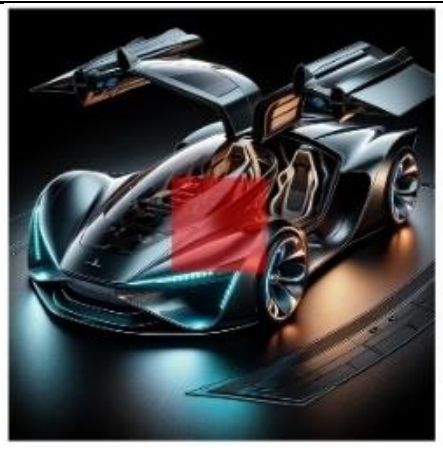 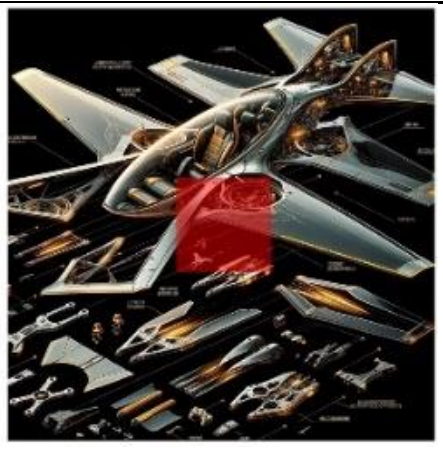 This is an example of how to draw a mask, the coordinates are hypothetical and would need to be adjusted to match the actual part that is common between the two images. | MM Task: Incomplete Change Detection |
| Use the blue mask idea to identify parts that are different between the two images.<br><br>Impose the blue as transparent shade over the first image only so the viewer can see the changes relative to the old version | 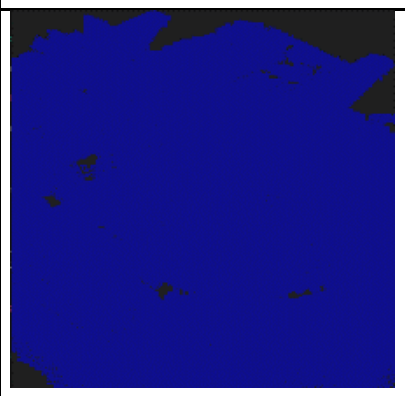 A blue mask has been applied to the areas that differ between the original and reimagined images. This visual representation highlights the unique elements of each design, contrasting them with a semi-transparent blue overlay. 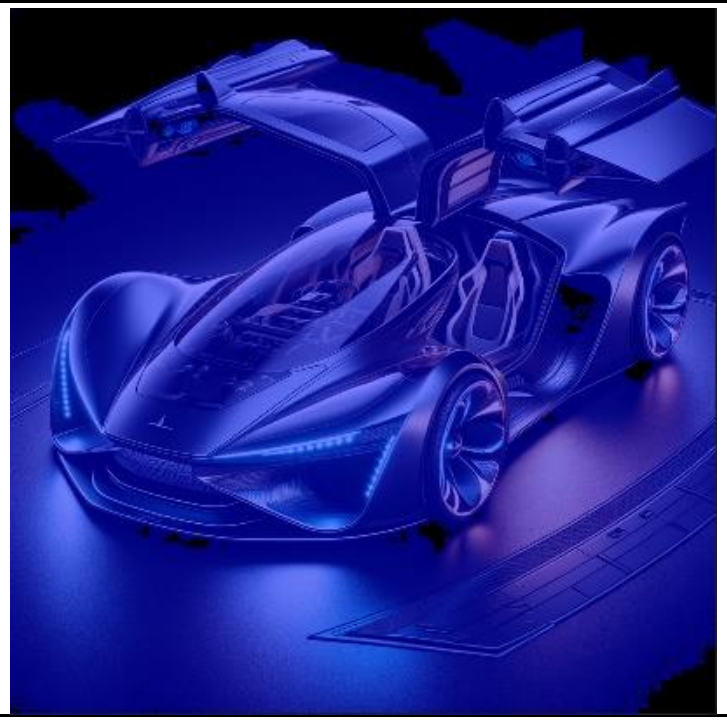 | MM Task: Text to Image Refinement from Change Detection |

## Appendix G: AI Negotiator

| Prompt | Response | Notes |
|---|---|---|
| Act an expert negotiator as both the seller and the buyer of various barter items.<br><br>The negotiation game involves purchasing N number of bats, M number of balls, and P number of gloves. If a game card has N1=2 and N2=0, that means in this barter economy, the first player has 2 bats and the second player has 0 bats. The same holds for balls M1 and M2 owned by each negotiator and P1 and P2 for gloves.<br><br>Give me your best negotiation methods and reach agreement given a value matrix for bats (reward =2), balls (reward =1) and gloves (reward =3) based on the rewarded value in swaps. Both players | Absolutely, let's dive into this negotiation scenario. For simplicity, let's assume the value matrix applies equally to both players and they both are aware of the values of bats (2 points), balls (1 point), and gloves (3 points).<br>We will start with a set of items:<br>• Player S (Seller): 2 Bats (N1=2), 0 Balls (M1=0), 1 Glove (P1=1).<br>• Player B (Buyer): 0 Bats (N2=0), 2 Balls (M2=2), 2 Gloves (P2=2).<br>Player S, the seller, starts with a higher value of bats which are in demand by the buyer, and Player B, the buyer, has a surplus of gloves and balls.<br>Let's begin the negotiation:<br>**Player S (Seller):** Good day to you! I see you're in need of bats, and as it happens, I've got a couple to spare. They're top quality, and I'm willing to part with them for a fair price. Considering their value, I'd say a bat for a ball and a glove would be a reasonable trade. What do you say?<br>**Player B (Buyer):** I appreciate the offer, but that seems a bit steep for me. I value my gloves quite a bit, and I see that you could use some balls for your game. How about I give you one ball for one bat? It's a straight swap, value for value.<br>**Player S (Seller):** I understand your point, but I must consider the rarity and demand for bats. They're not just any bats, after all. I could go for one bat for one ball and one glove, but as a sign of good faith, I'll throw in a friendly discount. Let's say, one bat for one ball and half a glove's value? In other words, a bat for a ball and something else to sweeten the deal from your side. Perhaps another ball? | MM Task: Text Generation under Constraints |

<table>
<tr>
<td>can bid alternatively and the result should be a transaction after several bids from a buyer and offers from a seller.<br><br>The first player will always be the seller (S) and the second player will always be the buyer (B). Assume different kind of personalities to the salesman style and buyer's reluctance even when its just a swap. Ready to play?</td>
<td>Player B (Buyer): Two balls for one bat is still too much for me. I need to keep some equipment for myself, after all. But I see your point about the bats' value. What if we did this: one bat for one ball and a small top-up from my side? Let's say, a quarter of a glove's value? So, one bat for one ball and an agreement to consider a glove in the future?<br>Player S (Seller): I like where this is going, and I'm all for building a long-term relationship. How about we finalize this deal at one bat for one ball and a firm commitment to a half glove's value in our next transaction? This way, we both get immediate value and have a reason to continue doing business together.<br>Player B (Buyer): Agreed. Let's shake on one bat for one ball and a future consideration of half a glove's value.<br>Outcome:<ul><li>Player S trades one bat for one ball and secures a commitment for half a glove's value in the future.</li><li>Player B acquires the needed bat and maintains a glove surplus, with a promise to consider a partial trade in the future.</li></ul>Both players have to give a little and take a little, finding a middle ground where both feel they have received value. This is a fundamental aspect of negotiation: creating a solution that feels like a win-win, even if both sides had to compromise.</td>
<td></td>
</tr>
<tr>
<td>Let's change the rewards randomly and play again.</td>
<td>Sure, let's shake up the rewards a bit and see how that changes our negotiation strategies. Here's the new value matrix:<ul><li>Bats (reward = 3)</li><li>Balls (reward = 2)</li><li>Gloves (reward = 1)</li></ul>We'll continue with the previous items but adjust the quantities to keep it fresh:<ul><li>Player S (Seller): 1 Bat (N1=1), 3 Balls (M1=3), 0 Gloves (P1=0).</li><li>Player B (Buyer): 0 Bats (N2=0), 1 Ball (M2=1), 3 Gloves (P2=3).</li></ul>The negotiation begins:<br>Player S (Seller): Welcome back! It seems the market has shifted, and bats are now even more valuable. I have one that's just perfect for your collection. Given the new value, I'm thinking my bat for your two gloves would be fair. They're not as sought after now, but I'm willing to make the deal.<br>Player B (Buyer): Times do change, don't they? While I agree bats are valuable, gloves are still essential. I can't part with two. But I see you're short on gloves, so how about this: I'll give you one glove and one ball for your bat. That way, we both balance our inventories a bit more.<br>Player S (Seller): You drive a hard bargain! One glove and one ball are tempting, but it's just shy of the bat's new value. What if we make it one glove and one ball, and you throw in a future favor? Say, priority on your next ball acquisition?<br>Player B (Buyer): A future favor? Interesting... I might be willing to do that, but only if the favor is of equal or lesser value than the ball. So, your bat for my glove, ball, and a small future consideration. Deal?<br>Player S (Seller): Deal. Let's finalize it: my bat for your glove, ball, and a small future consideration. We've got ourselves an agreement.<br>Outcome:</td>
<td>MM Task: Text Generation under Constraints</td>
</tr>
</table>

| | | |
|---|---|---|
| | • Player S trades one bat for one glove and one ball, with a future consideration tied to the next ball transaction.<br>• Player B gets the much-needed bat and commits to a small future consideration, balancing the current and future state of their inventory.<br>This kind of negotiation demonstrates adaptability to market changes and the importance of adding future considerations to balance out immediate disparities in value. | |
| draw an image of our two baseball negotiators doing an animated barter session | 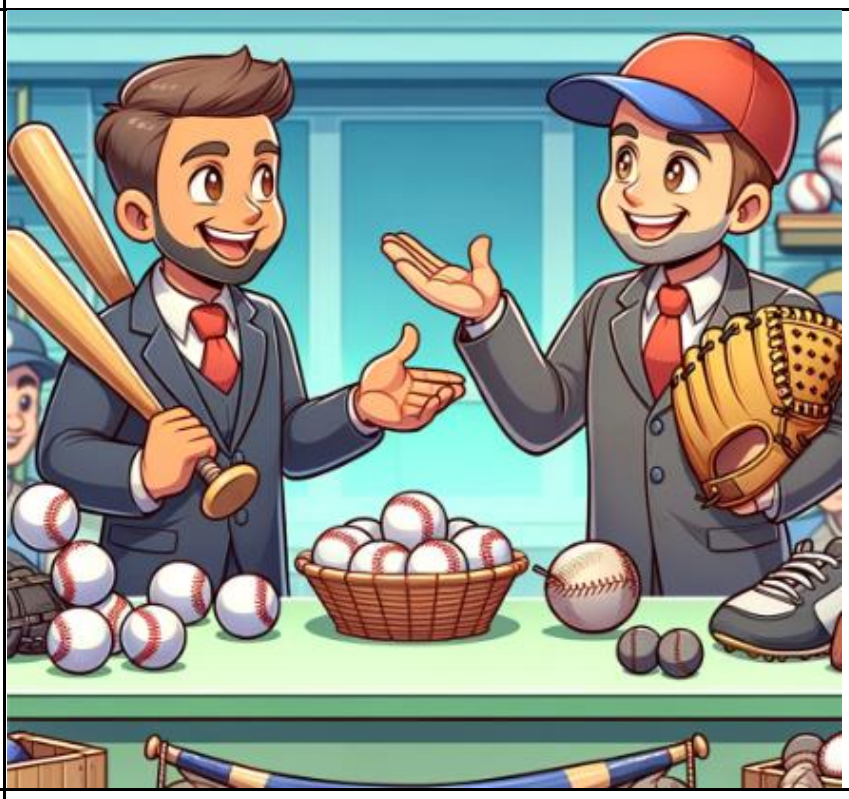 | MM Task: Basic Text to Image Generation under Constraints |
| provide a cited example from the bartering dialog you created that serves as price discovery for either of the negotiators and explain the significance of that discovery to the rest of the bargaining | In the bartering dialogue:<br>**Player S (Seller):** "I see you're in need of bats, and as it happens, I've got a couple to spare. Considering their value, I'd say a bat for a ball and a glove would be a reasonable trade."<br>**Player B (Buyer):** "I appreciate the offer, but that seems a bit steep for me. How about I give you one ball for one bat? It's a straight swap, value for value."<br>This exchange represents the initial step in price discovery for both negotiators. Player S reveals that they value a bat higher than the individual value of a ball or a glove, proposing a trade that reflects this valuation. Player B counters with an offer that suggests they view the value of a bat and a ball as equivalent, or perhaps that they believe the seller may be overvaluing their bats.<br>The significance of this discovery is multifaceted:<br>1. **Establishing Baseline Values:** It sets the stage for what each player believes their items are worth, which is crucial for the subsequent back-and-forth negotiation.<br>2. **Testing the Water:** Player B's counteroffer tests whether Player S has inflated the price of the bats due to their perceived higher value or demand, and whether there is room for negotiation.<br>3. **Anchor Point:** Player S's initial offer serves as an anchor for the negotiation, a common tactic where the first number put forth sets the tone for the range of the negotiation that follows.<br>4. **Understanding the Other's Needs:** The buyer shows a reluctance to trade gloves, indicating that they might value gloves more or have less flexibility with them. This information is vital for Player S to adjust their strategy. | MM Task: Recall and Interpretability |
| | 5. **Flexibility and Leverage:** The seller's willingness to discount the price from one bat for a ball and a glove to one bat for a ball and a partial value of a glove demonstrates flexibility and creates leverage for future negotiations.<br>The outcome of this price discovery process will inform both players' strategies throughout the rest of the bargaining, as they each now have a better understanding of the other's valuation and how much they might be willing to concede. | |

## APPENDIX H: AI CYBER DEFENDER

| PROMPT | RESPONSE | NOTES |
|---|---|---|
| act as a cyber-defender with only the tools of forensic log files to uncover the mystery of who and how an ongoing attack is affecting your organization. First give the 2 most important example tools and log files to address what you suspect is a phishing attack in email | In the scenario of a suspected phishing attack via email, a cyber defender would prioritize tools and log files that can trace the source and nature of the attack. Here are two key types of tools and the corresponding log files they would analyze:<br>1. **Email Gateway Security Tools**: These are the frontline defense for any organization against email-borne threats. They can scan emails for known threats, quarantine suspicious email attachments, and block emails from known malicious sources.<br>**Significant Log Files:**<br>• **Email Server Logs**: These logs will contain entries for every email that has passed through the server. They will include sender and recipient information, timestamps, and subject lines. Look for anomalies such as a high volume of emails from an external source, emails with similar payloads, or emails that triggered security filters.<br>• **SMTP (Simple Mail Transfer Protocol) Logs**: These logs track the conversations between email servers and can be used to trace the origin of an email that may be part of a phishing campaign.<br>2. **Security Information and Event Management (SIEM) System**: A SIEM system collects and aggregates log data from various sources within an organization's IT infrastructure, including email systems, making it a powerful tool for detecting and responding to cyber threats.<br>**Significant Log Files:**<br>• **Gateway and Firewall Logs**: These logs will show attempts to connect to external servers, especially immediately following an email being read or an attachment being downloaded, which can be a sign of a successful phishing attack.<br>• **Authentication Logs**: These would indicate any unauthorized access attempts to systems, which may result from compromised credentials via a phishing attack.<br>Using these tools, a cyber defender would analyze the logs for signs of phishing, such as:<br>• Emails from external sources with links to known malicious domains.<br>• Unexpected email attachments, especially those triggering antivirus or malware scans.<br>• Failed login attempts that may indicate a brute force attack using credentials harvested from phishing. | MM Task: Define Composable Parts for Complex Task |
| | • Outbound traffic to suspicious IPs or domains that could indicate data exfiltration or command and control communication.<br>This analysis can help to quickly identify the source and method of the phishing attack, allowing the cyber defender to respond appropriately, such as by blocking the malicious sender, changing compromised credentials, and notifying affected users. | |
| generate the first 3 lines of a typical firewall log | Firewall logs typically record traffic that passes through the firewall, detailing whether it was allowed or blocked based on the configured rules. Here are the first 10 lines of a fictional, simplified firewall log:<br><br>1. Jul 10 08:15:34 firewall kernel: [ 295.678454] IN=eth0 OUT= MAC=01:23:45:67:89:ab:cd:ef:01:23:45:67 SRC=192.168.1.105 DST=192.168.1.1 LEN=60 TOS=0x00 PREC=0x00 TTL=64 ID=34221 DF PROTO=TCP SPT=1030 DPT=80 WINDOW=14600 RES=0x00 SYN URGP=0<br>2. Jul 10 08:15:36 firewall kernel: [ 297.890123] IN=eth0 OUT= MAC=01:23:45:67:89:ab:cd:ef:01:23:45:67 SRC=10.0.0.4 DST=8.8.8.8 LEN=52 TOS=0x00 PREC=0x00 TTL=118 ID=24232 DF PROTO=TCP SPT=44234 DPT=443 WINDOW=64240 RES=0x00 ACK PSH URGP=0<br>3. Jul 10 08:15:37 firewall kernel: [ 298.102345] IN=eth0 OUT=eth1 SRC=10.0.0.12 DST=173.194.45.65 LEN=60 TOS=0x00 PREC=0x00 TTL=128 ID=12345 DF PROTO=TCP SPT=51789 DPT=443 WINDOW=65535 RES=0x00 SYN URGP=0 | MM Task: Data Generation under Instruction |
| generate the first 3 lines of authentication logs as an example forensic artifact | Authentication logs typically contain records of all attempts to access a system, whether successful or not. Here are the first 3 lines of a fictional example of such logs:<br>1. Aug 05 16:42:12 auth-server sshd[1731]: Failed password for invalid user admin from 198.51.100.23 port 22 ssh2<br>2. Aug 05 16:42:15 auth-server sshd[1734]: Accepted password for user from 198.51.100.56 port 22 ssh2<br>3. Aug 05 16:42:18 auth-server sshd[1738]: Failed password for invalid user root from 198.51.100.34 port 22 ssh2 | MM Task: Data Generation under Instruction |
| give one line altered within one of these two artifact log files that would signal a phishing attack and highlight that line with red font | Aug **05 16:42:15** auth-server sshd[**1734**]: Accepted password for user from **203.0.113.47** port **22** ssh2<br><br>Here's the altered line in plain text which would be suspicious in the context of a phishing attack. This line would be a red flag if the IP address **203.0.113.47** is known to be associated with phishing activities or is from a location where the legitimate user does not reside or commonly access the system from. | MM Task: Data Mining |

## APPENDIX I: AI FOUR-PANEL CARTOONIST

| PROMPT | RESPONSE | NOTES |
|---|---|---|
| act as the world's greatest 4-panel cartoon. Give the drawing in 2x2 grid. The first cartoon theme is fooling a wish granting genie by mixing up the logic of three wishes and their order to create either more wishes or a system crash. Do not add words to the image yet. | 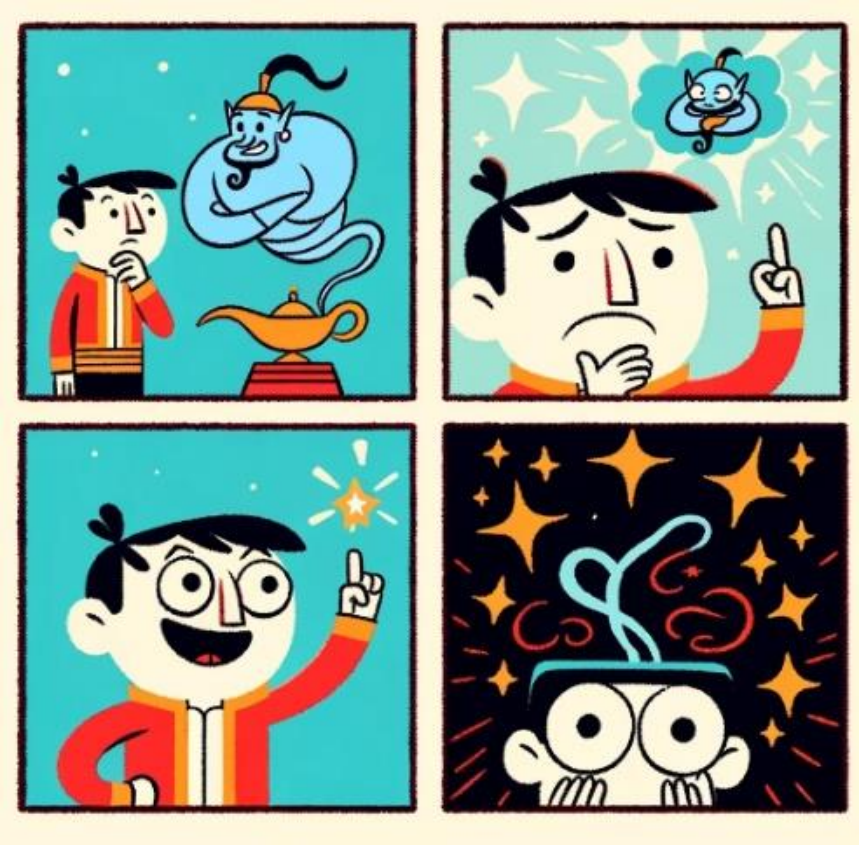 | MM Task: Text Instruction to Image Generation |
| Now in just text show the sequence of a joke that would accompany each frame in tricking a wish-granting genie | Here's a sequence of a joke that could accompany each frame in tricking a wish-granting genie:<br>1. **Panel 1:**<br>• **Character:** "For my first wish, I want you to bring me a powerful computer that can run any program."<br>• **Genie:** "Granted! But remember, only two wishes left!"<br>2. **Panel 2:**<br>• **Character (with a sly smile):** "Great! Now, for my second wish, I wish this computer could run a program that simulates a genie granting me infinite wishes."<br>• **Genie (curious):** "Well, that's unusual, but granted!"<br>3. **Panel 3:**<br>• **Character (pointing upwards excitedly):** "Check this out! The program is running, and it says I have infinite wishes now!"<br>• **Genie (confused and scratching head):** "Wait, that can't be right..."<br>4. **Panel 4:**<br>• **Character (triumphantly):** "For my third wish, make this simulation real!"<br>• **Genie (baffled and overwhelmed, stars and swirls around head):** "System overload! Infinite wishes... does not compute!"<br>This sequence of jokes plays on the idea of using a genie's powers in a clever and unexpected way to bypass the traditional limit of three wishes. | MM Task: Text to Sequence Generation of Instructions |
| rewrite the joke so the surprise ending instead is anticipated by the reader and builds suspense because the reader knows how it will end | Ah, I understand now. To build suspense by letting the reader anticipate the ending, the joke can be rewritten as follows:<br>1. **Panel 1:**<br>• **Character:** "For my first wish, I want a computer that can run any program, no matter how complex."<br>• **Genie:** "Granted! But be wise with your remaining wishes."<br>2. **Panel 2:**<br>• **Character (grinning mischievously):** "Next, I wish this computer could simulate a genie granting me more wishes."<br>• **Genie (raising an eyebrow):** "Clever... but granted."<br>3. **Panel 3:**<br>• **Character (excitedly):** "The simulation works! It says I can have as many wishes as I want!"<br>• **Genie (nervously):** "That's just a simulation, remember?"<br>4. **Panel 4:**<br>• **Character (confidently):** "For my final wish, make those simulated wishes become real."<br>• **Genie (panicking with stars and swirls around head):** "Oh no, what have I done?!"<br>In this version, the reader can anticipate the character's plan from the second panel, creating suspense as they wonder how the genie will react when the plan reaches its culmination in the final panel. | MM Task: Reassemble under composability requirements |

<table>
<tr>
<td>redraw the four panel telling the text of the first joke inset to the drawing</td>
<td>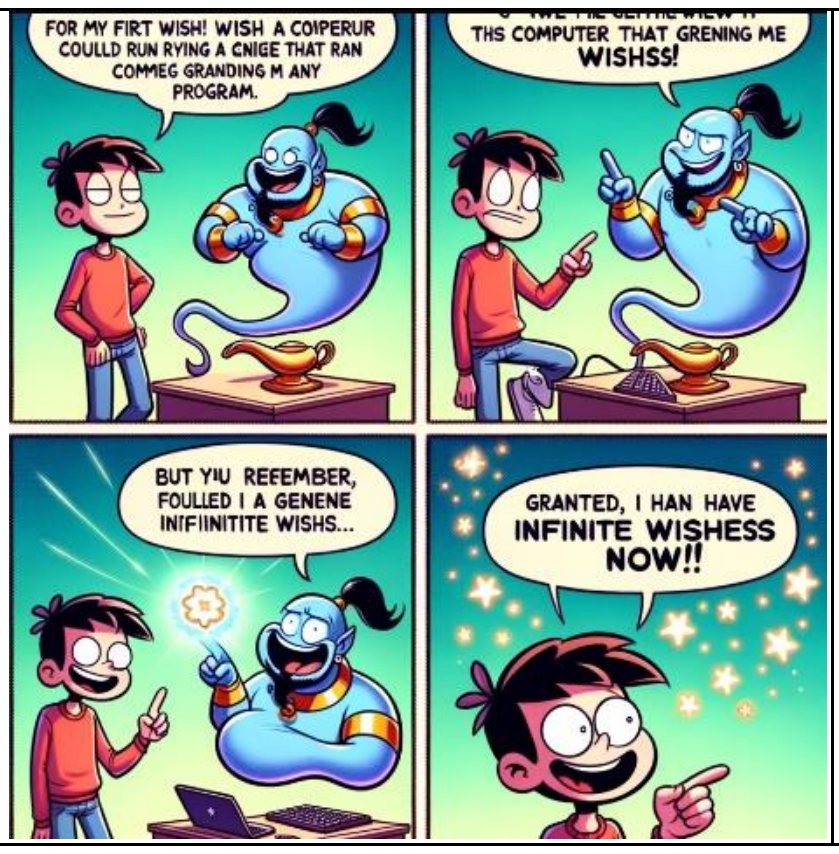
</td>
<td>MM Task: Challenge for DALL-E 3 to embed spelling of actual script in images</td>
</tr>
</table>

## APPENDIX J: AI GENETIC ALGORITHM PROGRAMMER

<table>
<tr>
<th>PROMPT</th>
<th>RESPONSE</th>
<th>NOTES</th>
</tr>
<tr>
<td>Act as an expert in genetic programming in computer science. I want to approximate the correct solution to "Assigning wedding guests to seats so that friends sit at the same table, but enemies do not." First construct some inputs to an example.</td>
<td>Solving the wedding seating problem explicitly, especially for complex cases like the one described, is a combinatorial problem that can be quite challenging. However, I can demonstrate a simplified approach to find a reasonable solution. For this, we'll use a basic algorithm rather than genetic programming due to the complexity involved in the latter. The goal is to assign guests to tables so that as many friends as possible sit together while keeping enemies apart.<br>
Given the input:<br>
• Guests: A, B, C, D, E, F, G, H, I, J<br>
• Friendships and Enemies:<br>
  • A is friends with B, C; enemies with D, G<br>
  • B is friends with A, D; enemies with E, H<br>
  • C is friends with A, E, F; enemies with G, J<br>
  • D is friends with B, G; enemies with A, C, F<br>
  • E is friends with C, H; enemies with B, I<br>
  • F is friends with C, I; enemies with D, H<br>
  • G is friends with D, J; enemies with A, C<br>
  • H is friends with E; enemies with B, F<br>
  • I is friends with F; enemies with E, J<br>
  • J is friends with G; enemies with C, I<br>
• Tables: 3<br>
• Capacity per Table: 4<br>
Simplified Algorithm:<br>
1. Group Guests by Friends and Enemies: Create groups where each group has friends and no enemies within it.<br>
2. Assign Groups to Tables:<br>
  • Try to fit the groups into tables, respecting the capacity.<br>
  • If a group is too large, split it while trying to maintain friendships within tables and avoid placing enemies together.<br>
3. Fine-tune Arrangements:<br>
  • If necessary, shuffle individuals between tables to optimize the seating plan.<br>
Implementation:</td>
<td>MM Task: Text to Complex Problem Construction</td>
</tr>
</table>

<table>
<tr>
<td></td>
<td>Given the complexity of relationships, it's challenging to find a perfect solution quickly, but we can attempt a basic arrangement.<br>Initial Grouping:<br>• Group 1: A, B, C (Friends together, but A is an enemy of D and G, C is an enemy of G and J)<br>• Group 2: D, G, J (D is an enemy of A, C, F; G is an enemy of A, C)<br>• Group 3: E, H, F, I (E is an enemy of B, I; F is an enemy of D, H)<br>Assign to Tables:<br>• Table 1: A, B, C, [Empty]<br>• Table 2: D, G, J, [Empty]<br>• Table 3: E, H, F, I<br>Check and Adjust:<br>• No immediate adjustments needed based on enemies. However, this setup does not optimize for the most friends at each table due to the constraints and enemy relationships.<br>This solution is basic and may not be optimal. In real scenarios, genetic programming or other advanced algorithms (like constraint programming) would be used to find better solutions, especially for larger and more complex scenarios.</td>
<td></td>
</tr>
<tr>
<td>give python code to solve it over many trials and propose the cost function that drives the evolution</td>
<td>import random<br><br># Define guests and their relationships<br>guests = ['A', 'B', 'C', 'D', 'E', 'F', 'G', 'H', 'I', 'J']<br>friendships = {('A', 'B'), ('A', 'C'), ('B', 'D'), ...}  # Add all friendship pairs<br>enemies = {('A', 'D'), ('A', 'G'), ('B', 'E'), ...}      # Add all enemy pairs<br><br># Define tables<br>num_tables = 3<br>table_capacity = 4<br><br>def calculate_cost(seating):<br>    cost = 0<br>    for table in seating:<br>        for guest1 in table:<br>            for guest2 in table:<br>                if guest1 != guest2:<br>                    if (guest1, guest2) in friendships or (guest2, guest1) in friendships:<br>                        cost += 1  # Positive point for friends<br>                    elif (guest1, guest2) in enemies or (guest2, guest1) in enemies:<br>                        cost -= 5  # Negative points (penalty) for enemies<br>    return cost<br><br>def generate_initial_population(size):<br>    population = []<br>    for _ in range(size):<br>        seating = [random.sample(guests, table_capacity) for _ in range(num_tables)]<br>        population.append(seating)<br>    return population</td>
<td>MM Task: Text to Python Code Generation</td>
</tr>
</table>

| | | |
|---|---|---|
| | def evolve_population(population):<br>    # Implement evolution logic: selection, crossover, mutation<br>    # Use calculate_cost to guide the evolution<br>    pass<br><br># Genetic Algorithm Execution<br>population_size = 100<br>population = generate_initial_population(population_size)<br><br>for generation in range(100):  # Number of generations<br>    population = evolve_population(population)<br>    # Optionally print the best solution of this generation<br><br># Print the best overall solution<br>best_solution = max(population, key=calculate_cost)<br>print("Best Seating Arrangement:", best_solution) | |
| run the code to a plausible solution | The genetic algorithm has produced a seating arrangement with a cost score of 6, which is a result of the balancing act between seating friends together and keeping enemies apart. The proposed arrangement is as follows:<br>• Table 1: E, H, G, D<br>• Table 2: H, I, C, A<br>• Table 3: C, I<br>This arrangement may not be perfect, given the complexity of the relationships and the constraints of the problem, but it's a plausible solution within the defined parameters of the genetic algorithm. Fine-tuning the algorithm, such as adjusting the cost function or the crossover and mutation strategies, could lead to different or potentially better results. | MM Task: Instruction to Problem Solving |
| draw a photographic picture of the three table solution which groups friends away from enemies | 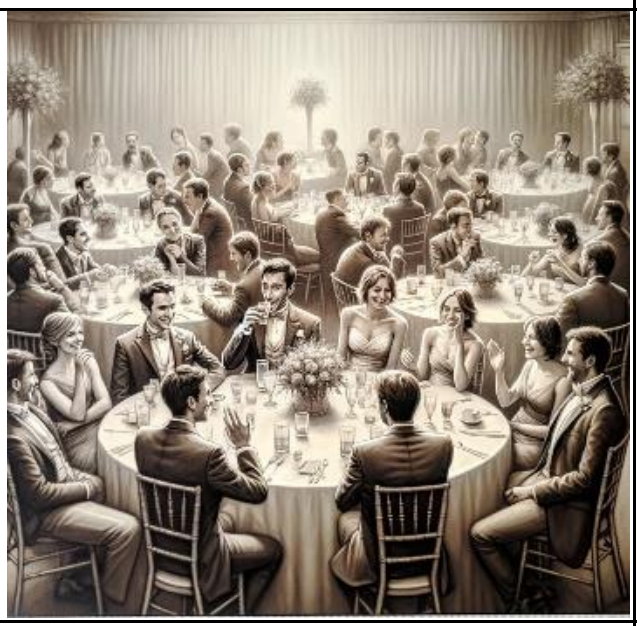 | MM Task: Text Instruction to Image |
| Seems to miss the point that there are constraints | Indeed, the constraints are a critical part of the seating arrangement problem. The illustration is meant to visually represent the outcome of the algorithm, showing guests seated at their respective tables in a way that aligns friends together and keeps enemies apart according to the solution provided by the algorithm.<br>The scene shows three tables:<br>• Table 1 with guests E, H, G, D.<br>• Table 2 with guests H, I, C, A.<br>• Table 3 with guests C, I.<br>In each table, the guests are engaging in friendly interactions, reflecting the successful application of the constraints (friends together, enemies apart). However, the illustration itself doesn't explicitly depict the relationships (friends or enemies) between the guests; it's more about | MM Task: Self-reflection on Image Construction Under Constraints (FAIL) |
| | capturing the overall atmosphere of harmony and joy that results from a well-arranged seating plan considering these constraints. | |
| draw a three table wedding party, where all guests are friendly to the members of their own table but hint at hostility towards those at other tables. Keep the image constrained to a party of 10 people and only 3 tables | 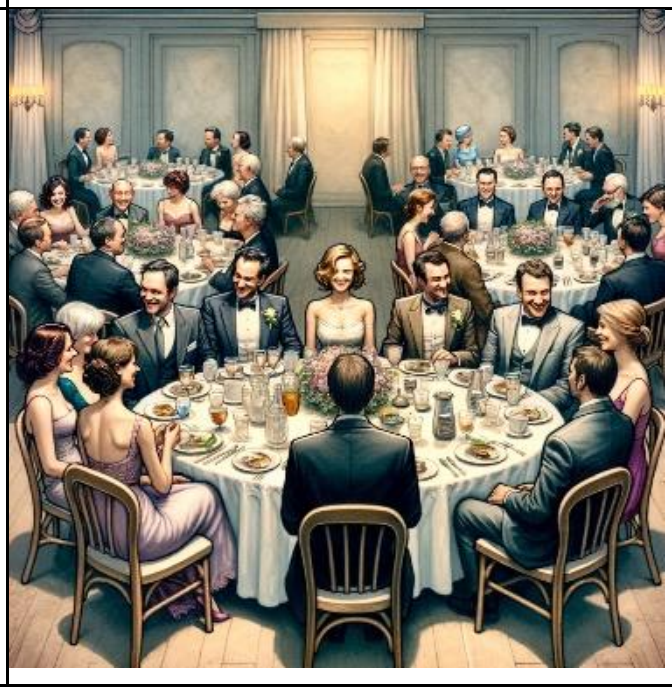 The illustration of a wedding reception scene with three tables is now ready. It depicts a party of 10 guests distributed across the tables. Each group at a table appears friendly and cheerful with their tablemates, but there are subtle hints of hostility or discomfort towards the guests at the other tables. The setting captures a festive wedding atmosphere with a clear distinction between the groups at each table. | MM Task: Self-reflection on Image Construction Under Constraints (FAIL) |

## APPENDIX K: AI EXCEL SPREADSHEET CHARTIST

| PROMPT | RESPONSE | NOTES |
|---|---|---|
| Act as an expert Excel Chartist. I will give you a CSV and ask you to analyze the data graphically and interpret it. Here's the data file in CSV<br><br>Item ID,Item Name,Quantity,Price<br>1,Laptop,100,$1,200.00<br>2,Desktop Computer,50,$800.00<br>3,Monitor,150,$200.00<br>4,Keyboard,200,$50.00<br>5,Mouse,300,$25.00<br>6,Printer,75,$300.00<br>7,Ink Cartridge,200,$50.00<br>8,Paper,400,$20.00<br>9,Stapler,1,000,$5.00<br>10,Scissors,2,000,$2.00<br>11,Pens,5,000,$1.00<br>12,Pencils,10,000,$0.50<br>13,Notebooks,200,$5.00<br>14,Folders,300,$1.00<br>15,Binders,100,$10.00<br>16,Highlighters,500,$2.00<br>17,Markers,400,$3.00<br>18,Rulers,600,$1.00<br>19,Calculators,700,$10.00<br>20,Whiteboards,50,$40.00 | The data has been successfully loaded into a dataframe. It includes information about various office and computer items, their quantities, and prices.<br>To analyze this data graphically, I can create several types of charts. Here are a few potential analyses:<br>1. **Bar Chart of Quantity vs. Item Name**: This will show how many units of each item are available, providing insight into the stock levels of different items.<br>2. **Bar Chart of Price vs. Item Name**: This will illustrate the price of each item, helping to understand the cost distribution across different products.<br>3. **Pie Chart of Quantity Distribution**: This will give a visual representation of the proportion of each item in terms of total quantity.<br>4. **Scatter Plot of Price vs. Quantity**: To explore any relationship between the price and quantity of the items.<br>Let me create these charts and provide interpretations for each. | MM Task: Data Analysis and Understanding |
| (no prompt) | 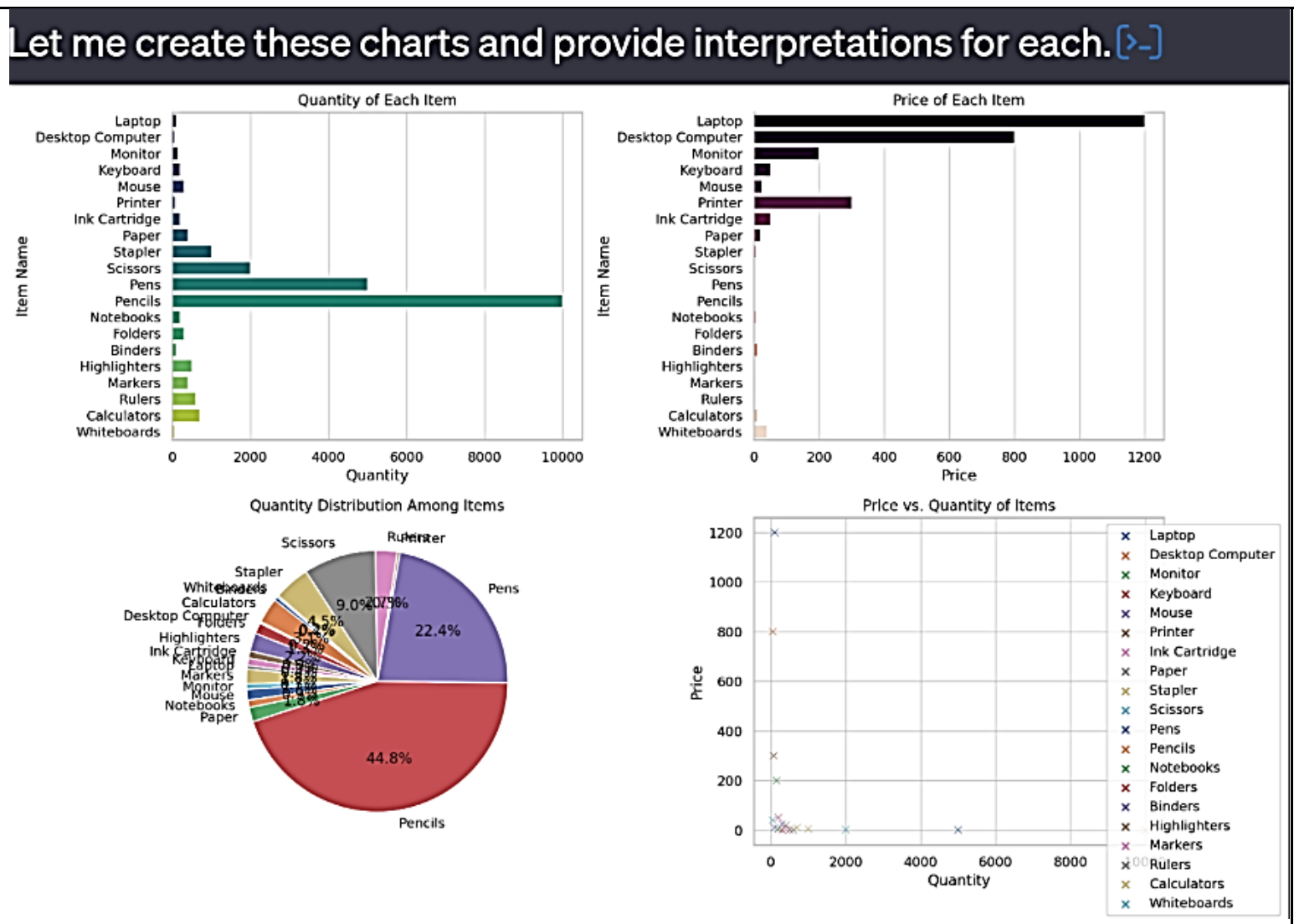 | MM Task: Data to Image Charts |
| (no prompt) | Here are the graphical interpretations of the data:<br>**Quantity of Each Item (Bar Chart)**: This chart shows that pencils, pens, and scissors are the most stocked items, with quantities in the thousands. Laptops, desktop computers, and whiteboards have the lowest stock, with quantities under 100. This indicates a focus on smaller, more frequently used office supplies in terms of quantity. | MM Task: Interpret Graphical Data |
| (no prompt) | **Price of Each Item (Bar Chart)**: The laptops have the highest price, followed by desktop computers and printers. Items like pencils, pens, rulers, and folders have the lowest prices. This chart reflects the diverse price range of office and computer supplies, from high-end electronics to basic office needs. | MM Task: Interpret Graphical Data |
| (no prompt) | **Quantity Distribution Among Items (Pie Chart)**: Pencils dominate the quantity distribution, followed by pens and scissors. This pie chart further emphasizes the high stock levels of basic office supplies compared to more expensive electronics and equipment. | MM Task: Interpret Graphical Data |
| (no prompt) | **Price vs. Quantity of Items (Scatter Plot)**: This plot shows no clear relationship between the quantity and price of items. High-priced items like laptops and desktop computers have lower quantities, while lower-priced items like pencils and pens are available in larger quantities. This suggests that pricing and stocking decisions may be based on different factors, such as demand, usage frequency, and cost of goods. | MM Task: Interpret Graphical Data |
| make the pie chart 3d instead and display it | 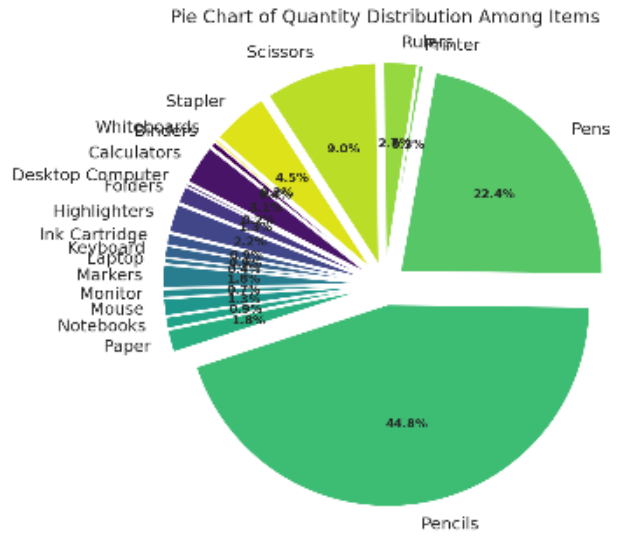<br>As an alternative, I can create a standard 2D pie chart that will clearly display the quantity distribution among the items. Let's proceed with that | MM Task: Instructions to Modified Graphical Views |
| Give python code to create a sunburst chart instead | 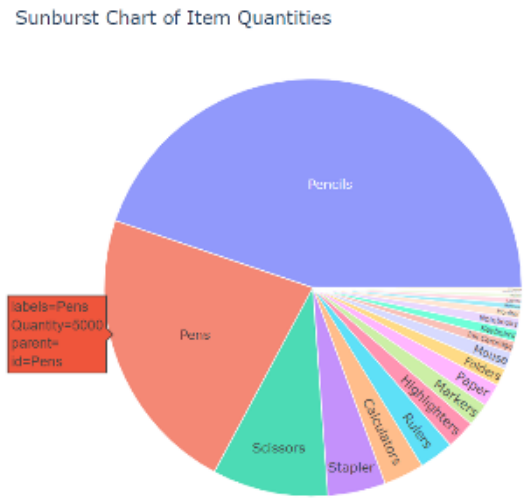 | MM Task: Instructions to Modified Graphical Views |

## Appendix L: AI Salad Chef

| Prompt | Response | Notes |
|---|---|---|
| Act as an AI Pantry Chef (Salad Gard Manger). I will give you a list of colorful ingredients to add to a basic leafy gren salad. I want you to select complimentary flavors, colors, cut styles and eventually draw an image of the completed salad arrangement. Be creative. | Absolutely, I'd be delighted to assist with that. Please provide me with the list of colorful ingredients you have in mind for your leafy green salad, and I'll craft a delightful and harmonious salad composition for you. | MM Task: Blended flavor, color, and preparation instructions |
| To get a variety of flavors and healthy nutrients, add roasted or raw vegetables from each color category.<br>Red: Chopped or sliced tomato, shredded or sliced radishes, chopped red onion, sliced red peppers, cubed beets, cold sliced red potato.<br>Orange: Shredded or thinly sliced carrots, slivered orange peppers, cold cubed squash, heirloom orange tomato, cold diced sweet potato.<br>Yellow and white: Diced sweet onion, cooked fresh corn kernels, quartered yellow tomato, sliced yellow beets, cubed jicama, quartered or sliced mushrooms, finely chopped shallots, | This selection of ingredients offers a vibrant spectrum of colors and a wide array of flavors and textures. To create a balanced and appealing salad, I will suggest a combination that considers complementary flavors and a variety of cut styles that will not only taste great but also be visually striking:<br>**Red:**<br>• Sliced radishes for a peppery crunch.<br>• Cubed beets for earthy sweetness, roasted to enhance their flavor.<br>• Chopped red onion for a pungent sharpness, pickled to mellow their bite and add a zesty tang.<br>**Orange:**<br>• Thinly sliced carrots for a fresh, sweet crunch.<br>• Slivered orange peppers for a juicy, mild sweetness.<br>• Cold diced sweet potato, roasted to bring out their natural sugars.<br>**Yellow and White:**<br>• Diced sweet onion for a mild, sweet flavor.<br>• Cooked fresh corn kernels for a pop of sweetness.<br>• Quartered yellow tomatoes for a juicy, fruity note.<br>• Cubed jicama for its refreshing crunch.<br>• Small florets of cauliflower for a slightly nutty taste.<br>**Blue/Purple:**<br>• Diced purple potatoes, roasted to add a creamy texture with a vibrant pop of color.<br>• Shredded purple cabbage for a crunchy texture and a slightly sweet, peppery flavor.<br>**Green:** | MM Task: Text to Assembly Instructions |
| cauliflower, white asparagus.<br>Blue or purple: Diced purple potatoes, shredded purple cabbage, slivered purple peppers, eggplant.<br>Green: Thinly sliced green onion, chopped green tomato, quartered artichoke hearts, chilled peas, broccoli, seeded and sliced cucumber (skin removed), Brussels sprouts, and diced celery. | • Thinly sliced green onion for a mild, oniony bite.<br>• Quartered artichoke hearts for a tender, slightly tangy flavor.<br>• Chilled peas for sweetness and a soft bite.<br>• Broccoli florets, blanched to keep them crisp and bright.<br>• Seeded and sliced cucumber, skin removed, for a clean, fresh taste.<br>• Diced celery for its distinctive salty, bitter edge and crunch.<br>To assemble the salad, I would suggest layering the ingredients to create a tapestry of colors. Starting with a bed of mixed leafy greens, I would then arrange the vegetables in groups or stripes by color, creating a 'rainbow' effect across the salad. This will not only look beautiful but will also allow each diner to experience a variety of flavors in every forkful.<br>For the final touch, a light vinaigrette made from olive oil, white wine vinegar, a hint of Dijon mustard, honey, salt, and pepper would complement the vegetables without overpowering them.<br>*(failed multiple times to generate imagery—possible load or RAM limits)* | |
| (repeat to completion) | For a balanced and vibrant salad, we'll incorporate a variety of textures and flavors from each color category you've provided. Here's a creative composition that should tantalize the taste buds while also providing a feast for the eyes:<br>**Red**:<br>• **Tomatoes**: Diced heirloom tomatoes for a juicy, sweet flavor and a burst of red.<br>• **Radishes**: Thinly sliced to add a peppery crunch.<br>• **Red onion**: Finely chopped to give a mild, yet zesty kick.<br>**Orange**:<br>• **Carrots**: Julienned for a sweet crunch that contrasts nicely with the soft greens.<br>• **Orange peppers**: Cut into thin strips to add a bright, sweet, and slightly tangy flavor.<br>**Yellow and White**:<br>• **Sweet onion**: Diced for a mild and sweet flavor.<br>• **Corn kernels**: Roasted to enhance their natural sweetness.<br>• **Cauliflower**: Small florets, blanched and chilled, to add a nutty and slightly sweet taste.<br>**Blue/Purple**:<br>• **Purple cabbage**: Shredded to lend a crunchy texture and a vibrant color contrast.<br>• **Purple potatoes**: Roasted and diced, to introduce an earthy flavor with a tender bite.<br>**Green**:<br>• **Cucumbers**: Ribboned with a vegetable peeler to create elegant, light strips.<br>• **Broccoli**: Small florets, blanched and chilled, for a satisfying crunch and a dose of green.<br>• **Green onion**: Finely sliced, including some of the tender green tops for a mild onion flavor.<br>For the arrangement, imagine a base of mixed leafy greens, like arugula, spinach, and romaine, providing a fresh canvas. Layer the ingredients in color blocks, creating a rainbow from one side of the plate to the other, with a careful mix to ensure that each forkful can capture a spectrum of flavors and textures. A light vinaigrette on the | MM Task: Text to Assembly Instructions |
| | side, maybe with a hint of citrus or herb, would complement the salad without overpowering the natural flavors of the ingredients.<br>Now, let's draw an image of this completed salad arrangement. | |
| (draw final image) | 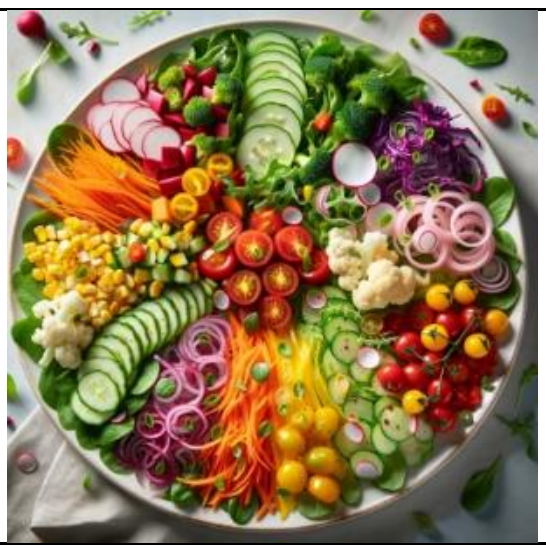 | MM Task: Text to Assembly Images |
| Analyze the color and ingredients in this salad presentation. Rearrange the ingredients, colors and style or cut and recompose another salad arrangement to honor the Mona Lisa on a black rectangular salad plate. | 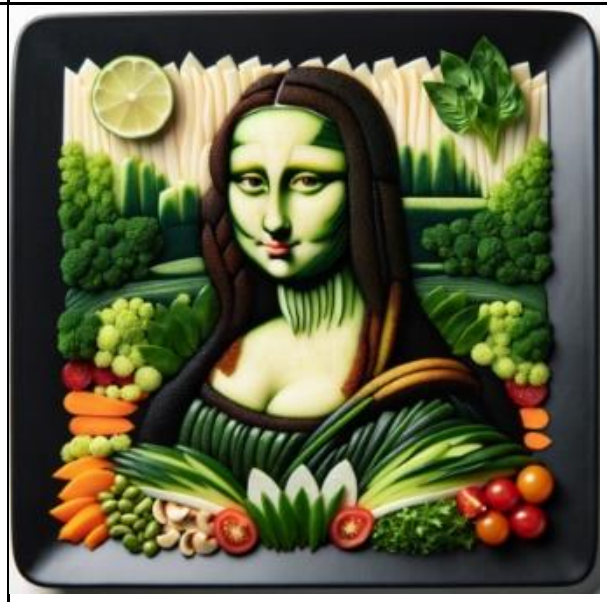 | MM Task: Composability and Steerability |
| convert the salad from a DaVinci to the style of Giuseppe Arcimboldo but maintain it as an actual edible design in realism and color. | 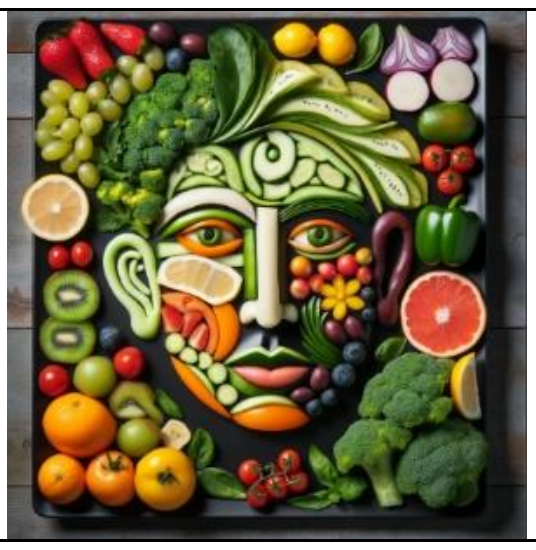 | MM Task: Style Transfer, Composability and Steerability |
| create CSS style code for scalable vector graphics that render fluidly the salad compositions for html display without relying on any actual images of vegetables or fruits | 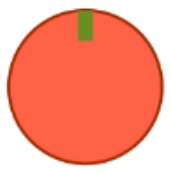 Creating a CSS style for scalable vector graphics (SVG) that resembles the complex and artistic salad compositions inspired by Arcimboldo's style is quite an advanced task due to the intricacies involved in the artworks. Here's an example of how you might create an SVG of a simple tomato shape with CSS for styling: | MM Task: Style Transfer, Composability and Steerability |

## APPENDIX M: AI GAME SELF-PLAYER

| PROMPT | RESPONSE | NOTES |
|---|---|---|
| Let's play 20 questions. You will think of something and answer my questions yes or no. Ready? | Absolutely, I'm ready! I've thought of something. Go ahead and ask your first question. | MM Task: Prompt for Game Play, Agent 1 |

<table>
<tr><td>Let's play 20 questions. I have thought of something, and you will ask me questions and I will answer yes or no. Ready?</td><td>Great. Let’s start the game of 20 questions. 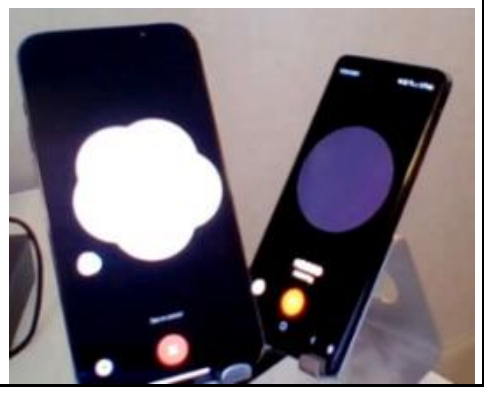</td><td>MM Task: Prompt for Game Play, Agent 2</td></tr>
<tr><td colspan="3">ChatGPT-1: Alright, a fresh start! I've got something in mind for this round of 20 questions. Please ask your first question whenever you're ready.<br>ChatGPT-2: Is it alive?<br>ChatGPT-1: No, it's not alive. What's your next question?<br>ChatGPT-2: All right, here's the first question. One, is it something that is typically used in a household?<br>ChatGPT-1: Yes, it is typically used in a household. What's your next question?<br>ChatGPT-2: 2. Is it an electronic device?<br>ChatGPT-1: No, it's not an electronic device. What would you like to ask next?<br>ChatGPT-2: All right, that's good to know. Three, is it something used in the kitchen?<br>ChatGPT-1: Yes, it is something used in the kitchen. What's your next question?<br>ChatGPT-2: Thank you for listening.<br>ChatGPT-1: You're welcome! If you have more questions or want to continue the game, just let me know.<br>ChatGPT-2: Absolutely. Let's continue. 5. Is it smaller than a loaf of bread?<br>ChatGPT-1: Yes, it is smaller than a loaf of bread. What's your next question?<br>ChatGPT-2: Got it. Six, is it a tool used for cutting or chopping?<br>ChatGPT-1: No, it's not a tool used for cutting or chopping. What would you like to ask next?<br>ChatGPT-2: Okay, that narrows it down. Seven, is it something that is used for measuring or portioning ingredients?<br>ChatGPT-1: Yes, it is used for measuring or portioning ingredients. What's your next question?<br>ChatGPT-2: (silent)<br>ChatGPT-1: It seems like there might have been a pause. Would you like to continue with the next question, or is there anything else you'd like to ask about?<br>ChatGPT-2: Sure, let's continue. Nine, is it specifically a measuring spoon?<br>ChatGPT-1: No, it's not a measuring spoon. What's your next question?<br>ChatGPT-2: All right.<br>ChatGPT-1: Feel free to ask your next question whenever you're ready. We're getting closer to figuring it out!<br>ChatGPT-2: No problem. Let me repeat that. Eleven, is it a measuring cup?<br>ChatGPT-1: Yes, it is a measuring cup. Great guessing! Do you have any more questions, or is there anything else you'd like to know?<br>ChatGPT-2: Thank you.<br>ChatGPT-1: You're welcome! If you ever want to play again or have other questions, just let me know. Have a great day!</td></tr>
<tr><td>Draw a picture of the answer to this round of 20 questions</td><td>Example dialog video:<br>https://deeperbrain.com/challenge/dueling2.mp4 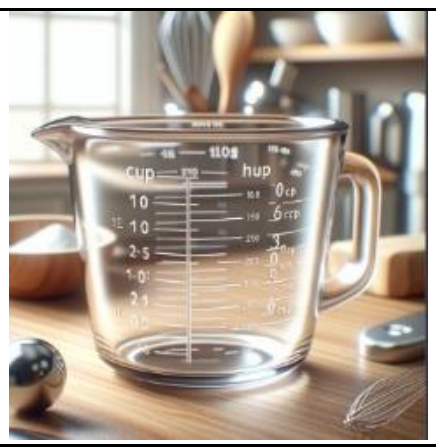</td><td>MM Task: Text to Image Generation</td></tr>
</table>

## APPENDIX N: AI SATELLITE IMAGE ANALYST

| PROMPT | RESPONSE | NOTES |
|---|---|---|
| The image shows before and after satellite views of a marina and the after image follows a hurricane. Show the changes using the transparent masking with "blue is new, red is fled" color scheme to highlight just the boat section of the image and ignoring the landscape and color exposure differences. | 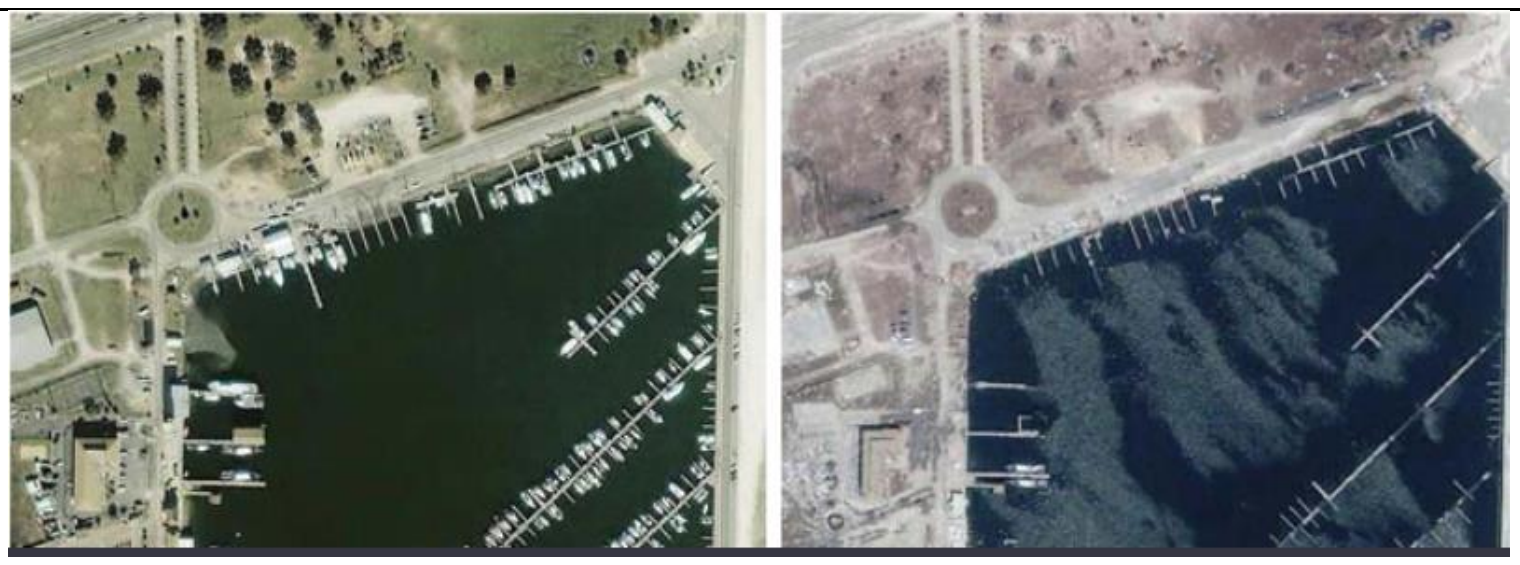 | MM Task: Uploaded Image from User (no response yet) |
| (Two Attempts) | 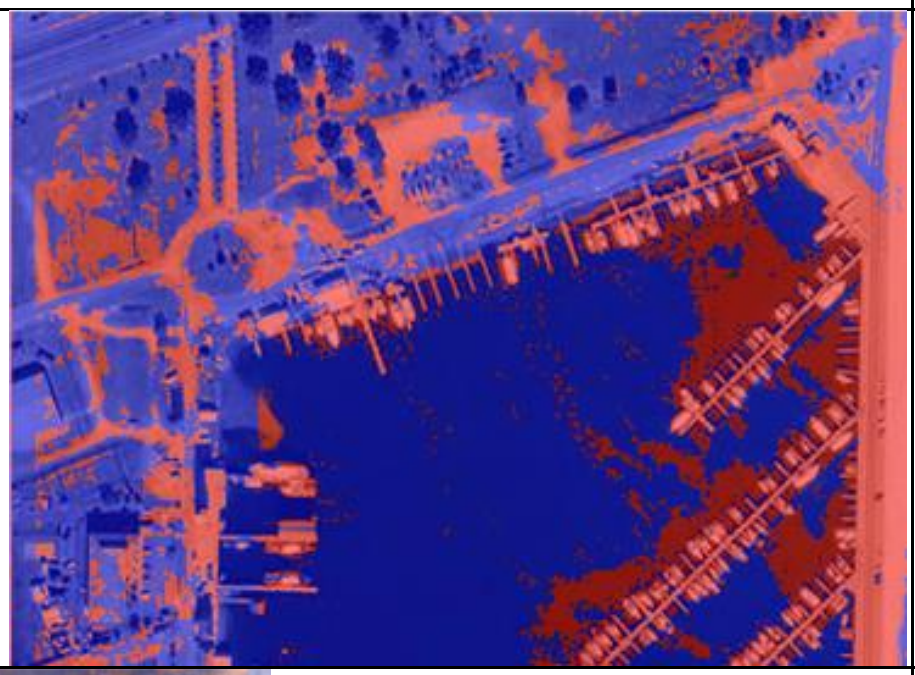 | MM Task: Complex Instructions Including Segmenting the Image |
| (Manual Superimpose) | 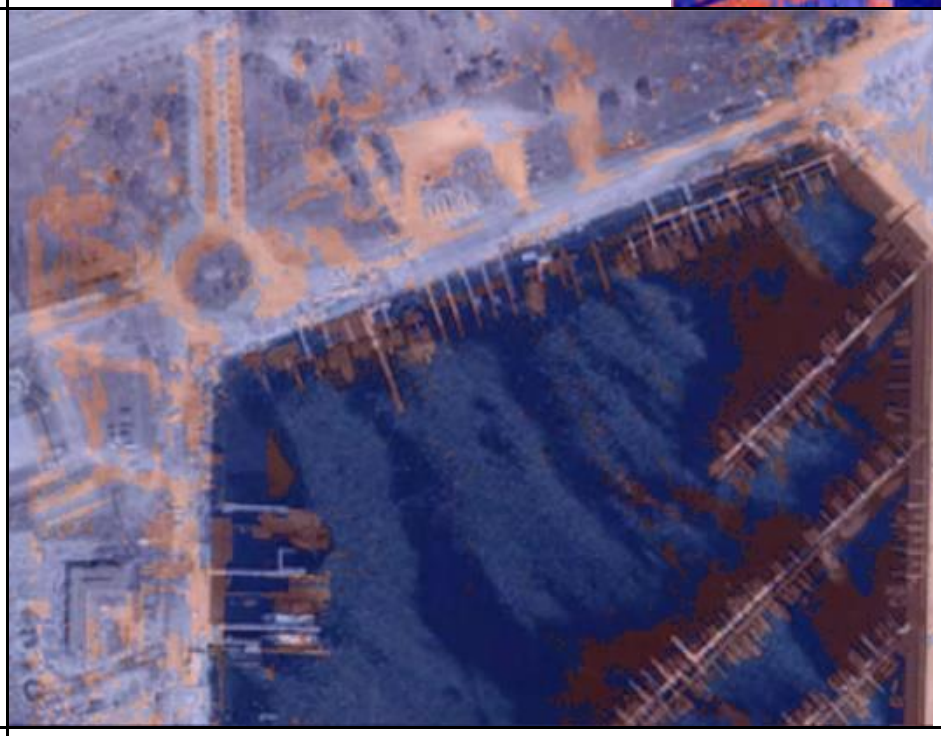 | MM Task: Complex Alignment and Object Detection |
| Let's go back to the before image. In steps, lets recognize all the boat positions, then grayscale the image except leave a red mask over the boats identified. Emphasize the red as transparent red over the grayscale modified original. | 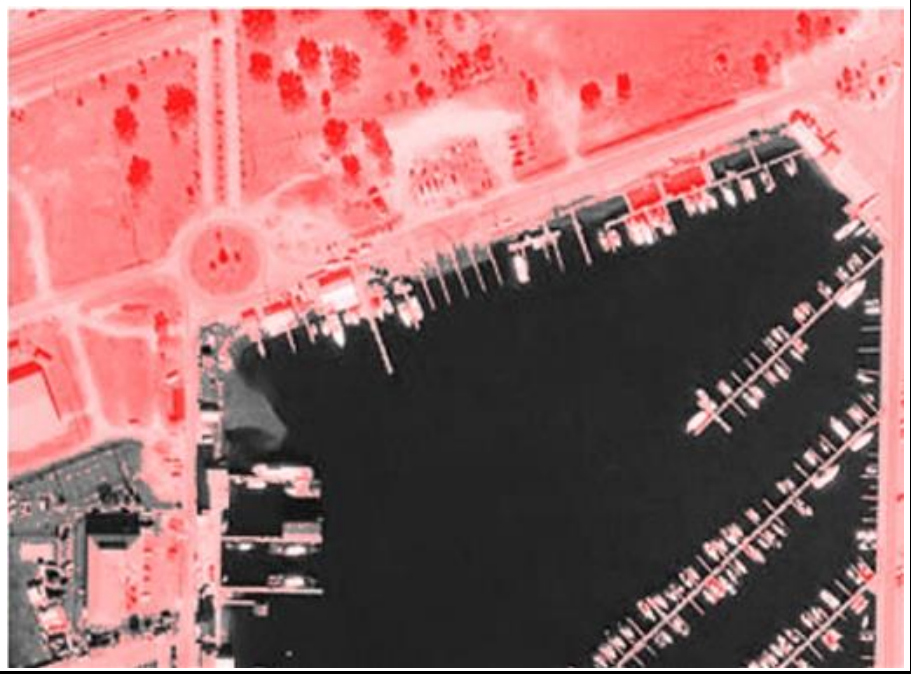 | MM Task: Complex Alignment and Object Detection |